\documentclass{article} 
\usepackage{iclr2025_conference,times}
\usepackage{booktabs}

\usepackage{amsmath,amsfonts,bm}

\def\eqref#1{equation~\ref{#1}}

\def\1{\bm{1}}

\DeclareMathAlphabet{\mathsfit}{\encodingdefault}{\sfdefault}{m}{sl}
\SetMathAlphabet{\mathsfit}{bold}{\encodingdefault}{\sfdefault}{bx}{n}

\usepackage{multirow, multicol}
\usepackage{hyperref}
\usepackage{url}
\usepackage{graphicx}
\usepackage{footnote}
\usepackage{courier}  

\usepackage{natbib}  
\usepackage{caption} 
\usepackage{color} 
\usepackage{placeins}
\usepackage{algorithm}
\usepackage{algorithmic}
\usepackage{dblfloatfix}
\usepackage{amsmath} 
\allowdisplaybreaks
\usepackage{subfig}

\title{A Hybrid Loss Framework for Decomposition-based Time Series Forecasting Methods: Balancing Global and Component Errors}

\author{Ronghui Han  \\
Depaertment of Mathematics\\
Sichuan University\\
China \\
\texttt{hronghui@stu.scu.edu.cn } \\
\And
Duanyu Feng\\
Depaertment of Computer Scienc \\
Sichuan University\\
China \\
\texttt{fengduanyuscu@stu.scu.edu.cn} \\
\AND
Hongyu Du \\
Department of Biostatistics and Informatics \\
University of Colorado Anschutz Medical Campus \\
USA\\
\texttt{hongyu.2.du@cuanschutz.edu}
\And
Hao Wang\thanks{Corresponding Author}\\
Depaertment of Mathematics \\
Sichuan University\\
China \\
\texttt{wangh@scu.edu.cn } \\
}

\iclrfinalcopy 
\begin{document}

\maketitle

\begin{abstract}
Accurate time series forecasting, predicting future values based on past data, is crucial for diverse industries. Many current time series methods decompose time series into multiple sub-series, applying different model architectures and training with an end-to-end overall loss for forecasting. However, this raises a question: does this overall loss prioritize the importance of critical sub-series within the decomposition for the better performance? To investigate this, we conduct a study on the impact of overall loss on existing time series methods with sequence decomposition. Our findings reveal that overall loss may introduce bias in model learning, hindering the learning of the prioritization of more significant sub-series and limiting the forecasting performance. To address this, we propose a hybrid loss framework combining the global and component losses. This framework introduces component losses for each sub-series alongside the original overall loss. It employs a dual min-max algorithm to dynamically adjust weights between the overall loss and component losses, and within component losses. This enables the model to achieve better performance of current time series methods by focusing on more critical sub-series while still maintaining a low overall loss. We integrate our loss framework into several time series methods and evaluate the performance on multiple datasets. Results show an average improvement of 0.5-2\% over existing methods without any modifications to the model architectures.

\end{abstract}

\section{Introduction}
Time series analysis is a powerful tool for understanding and forecasting sequential data points typically measured over time. It finds applications across various domains such as climate science~\citep{wu2023interpretable}, transportation~\citep{yin2021deep}, and energy~\citep{qian2019review}, where recognizing patterns and predicting future values are crucial.

Remarkably, deep learning methods have proven highly effective in time series forecasting by providing robust backbones/model architectures like Multilayer Perceptrons (MLPs)~\citep{zhang2022less,chen2023tsmixer}, Transformers~\citep{vaswani2017attention}, and even Large Language Models (LLMs)~\citep{jin2023time,openai2023gpt}, which are adept at learning complex patterns from large datasets~\citep{godahewa2021monash}. 
However, besides the improvement of the model architectures, most of these methods also rely on time series decomposition\citep{cleveland1990stl,QIAN2019939} to better capture various features. 
Among these, sliding-window decomposition is the most common method, which forms the basis of all model architectures discussed previously \citep{wu2021autoformer,zhou2022fedformer,nie2022time,zeng2023transformers}. It decomposes a raw time series into seasonal and trend sub-series, representing high-frequency feature (detailed changes) and low-frequency feature (overall trend changes), respectively~\citep{faltermeier2010sliding}.
However, although many methods utilize these sub-series, they still employ an end-to-end overall loss function. This loss function computes the difference between the final combined sub-series and the true series. This raises the question: Does optimizing this overall loss guarantee that the features of each sub-series are equally well-learned? Or, could an optimal overall loss fail to optimize the performance of the decomposition-based deep learning model?

To further investigate this, we conduct additional statistical analysis and case studies. Our statistical findings reveal that deep learning methods employing time series decomposition often exhibit significant discrepancies in losses across different sub-series on various datasets. Specifically, the loss on the seasonal sub-series is frequently one to two times smaller than the loss on the trend sub-series, which represents the overall movement of the time series. This disparity in losses suggests that the worse trend component may lead to substantial deviations in the overall trend of the forecast. We further illustrate this issue with a detailed case study.



To address this challenge, we propose a hybrid loss framework combining the global (the overall loss) and component errors (the sub-series losses). 
Inspired by the principles of distributionally robust optimization (DRO) \citep{wiesemann2014distributionally,namkoong2016stochastic,duchi2019variance}, we formulate this loss framework as a dual min-max problem. First, we construct a global min-max problem to balance the overall loss and the losses across all sub-series, ensuring that while minimizing the overall loss, the model also dynamically attends to the overall sub-series loss. Furthermore, recognizing that the overall sub-series loss is composed of individual sub-series losses, we formulate a second min-max problem to encourage the model to dynamically focus on potentially higher-loss components during training, thus prioritizing the optimization of critical components like the trend sub-series. We evaluate our loss framework on multiple datasets using existing model architectures and demonstrate an average performance improvement of 0.5-2\% without requiring any modifications to the underlying model structures.

In this paper, we make the following contributions: 
\begin{itemize}
    \item Our investigation reveals that the end-to-end overall loss function commonly used in deep learning for time series forecasting may not lead to optimal model performance. Sub-series critical to the overall forecasting might not be sufficiently optimized under the overall loss.
    \item We propose a novel hybrid loss framework that balances global and component errors to improve time series forecasting by dual min-max.
    \item The experiments demonstrate the effectiveness of our loss across diverse time series datasets, varying in both length and size, as well as across different models.
\end{itemize}

\section{Preliminary Experiments}
In this section, we explore a potential unifying issue among various deep learning approaches employing time series decomposition when trained under the current loss function. We illustrate this issue through statistical analysis of an experiment and by presenting several intuitive cases.

\textbf{Experiment Settings.} To investigate potential shortcomings of existing methods, we reproduce these decomposition-based deep learning methods and, beyond evaluating their overall performance, specifically analyze their performance on each decomposed sub-series. In our experiments,
\begin{itemize}
    \item For \underline{methods}, we select DLinear \citep{zeng2023transformers}, FEDformer \citep{zhou2022fedformer}, and PatchTST \citep{nie2022time} as representative methods. These methods all employ sliding-window-based time series decomposition (decompose to Seasonal sub-series and Trend sub-series), differing primarily in their backbone architectures: DLinear uses the MLP, while FEDformer and PatchTST utilize transformers. We employe the original loss function of these methods, which computes the Mean Squared Error (MSE) between the combined forecasting of the decomposed sub-series and the ground truth. Notably, these methods also represent the current state-of-the-art in time series forecasting in many benchmarks \citep{woo2022etsformer, wang2024timemixer}.
    \item  For \underline{datasets}, our experiments were conducted on four commonly used benchmark datasets: ETTh1, ETTh2 from ETTh \citep{haoyietal-informer-2021}, and ETTm1, ETTm2 from ETTm \citep{zhou2021informer}.  All datasets are split into training, validation, and testing sets with the 7:1:2 ratio.
    \item  For \underline{metrics}, performance is evaluated using the standard metrics of Mean Squared Error (MSE) and Mean Absolute Error (MAE).
\end{itemize}








We show the results of this experiment in Table \ref{tab:pre_ex}. With these results, we can find that \textit{\textbf{for deep learning methods employing sliding-window-based time series decomposition, significant discrepancies in forecasting performance across individual sub-series, when trained under the overall loss, may contribute to mainly inaccuracies in the final combined forecasting.}} Across the ETTh2, ETTm1, and ETTm2 datasets, the performance on the Trend sub-series is consistently 2 to 5 times worse than the performance on the Seasonal sub-series for all models. Conversely, on the ETTh1 dataset, the Seasonal sub-series performs approximately 2 times worse than the Trend sub-series. Furthermore, comparing the poorly predicted sub-series to the overall forecasting, it accounts for roughly 80\% of the overall error. This indicates that an overall loss may not ensure consistent predictive performance across individual sub-series for these decomposition-based methods, suggesting a biased learning towards certain components of sub-series. More importantly, this bias appears to be a major contributor to the overall forecasting error of these methods.

\begin{table}[htbp!]
  \centering
  \caption{ Multivariate time series forecasting results on four datasets with sliding-window-based deep learning methods. The results are based on the average of prediction lengths \{96, 192, 336, 720\} with input length 96. A lower MSE and MAE indicates better performance. The ``Global/Components" column indicates whether the reported results represent the overall forecasting performance or the performance on each individual decomposed sub-series.}
    \resizebox{0.95\textwidth}{!}{\begin{tabular}{cccccccccc}
    \toprule
          &       & \multicolumn{2}{c}{ETTh1\newline{}} & \multicolumn{2}{c}{ETTh2\newline{}} & \multicolumn{2}{c}{ETTm1} & \multicolumn{2}{c}{ETTm2} \\
    \midrule
    Models & Global/Componets & MSE   & MAE   & MSE   & MAE   & MSE   & MAE   & MSE   & MAE \\
    \midrule
    \multirow{3}[2]{*}{Dlinear} & Overall & 0.4588  & 0.4519  & 0.4981  & 0.4792  & 0.4061  & 0.4102  & 0.3102  & 0.3670  \\
          & Seasonal & 0.2965  & 0.3604  & 0.0888  & 0.2071  & 0.0969  & 0.2115  & 0.0486  & 0.1467  \\
          & Trend & 0.1716  & 0.3146  & 0.4144  & 0.4264  & 0.3192  & 0.3726  & 0.2661  & 0.3371  \\
    \midrule
    \multirow{3}[2]{*}{FEDformer} & Overall & 0.4394  & 0.4581  & 0.4429  & 0.4549  & 0.4441  & 0.4543  & 0.3031  & 0.3493  \\
          & Seasonal & 0.2793  & 0.3703  & 0.0866  & 0.2102  & 0.1010  & 0.2116  & 0.0446  & 0.1381  \\
          & Trend & 0.1678  & 0.3172  & 0.3551  & 0.4048  & 0.3249  & 0.3979  & 0.2595  & 0.3133  \\
    \midrule
    \multirow{3}[2]{*}{Patchtst} & Overall & 0.4506  & 0.4411  & 0.3658  & 0.3945  & 0.3838  & 0.3954  & 0.2821  & 0.3261  \\
          & Seasonal & 0.3031  & 0.3656  & 0.0835  & 0.1971  & 0.1319  & 0.2378  & 0.0490  & 0.1436  \\
          & Trend & 0.1566  & 0.2935  & 0.2710  & 0.3277  & 0.3198  & 0.3641  & 0.2328  & 0.2846  \\
    \bottomrule
    \end{tabular}
    }
  \label{tab:pre_ex}%
\end{table}%

To further explore the practical impact of this bias and provide a visual illustration, we conduct the case studies for each method across the different datasets, as shown in Figure \ref{fig:case}.\footnote{More cases can be found in Appendix \ref{More Showcases}.} We can find that \textit{\textbf{sub-series with larger losses, especially the Trend sub-series, does have a greater impact on the overall forecasting.}} In Figure \ref{fig:case} (a) and (b), the models accurately predict the Seasonal sub-series, but fail to capture the increasing trend in the Trend sub-series. This leads to a visually apparent underestimation of the overall forecasting compared to the ground truth. In contrast, for Figure \ref{fig:case} (c), the primary error occurs in the middle-early part, where both sub-series have significant errors. Although the Seasonal sub-series exhibits larger errors in the later part, the more accurate Trend sub-series forecasting results also can make a smaller overall error. Therefore, this further confirms that the overall loss may not effectively optimize for the sub-series that contribute significantly to the overall forecasting.

\begin{figure*}[htb!]
    \centering


    \subfloat[FEDformer on ETTh1.]{\includegraphics[scale=0.32]{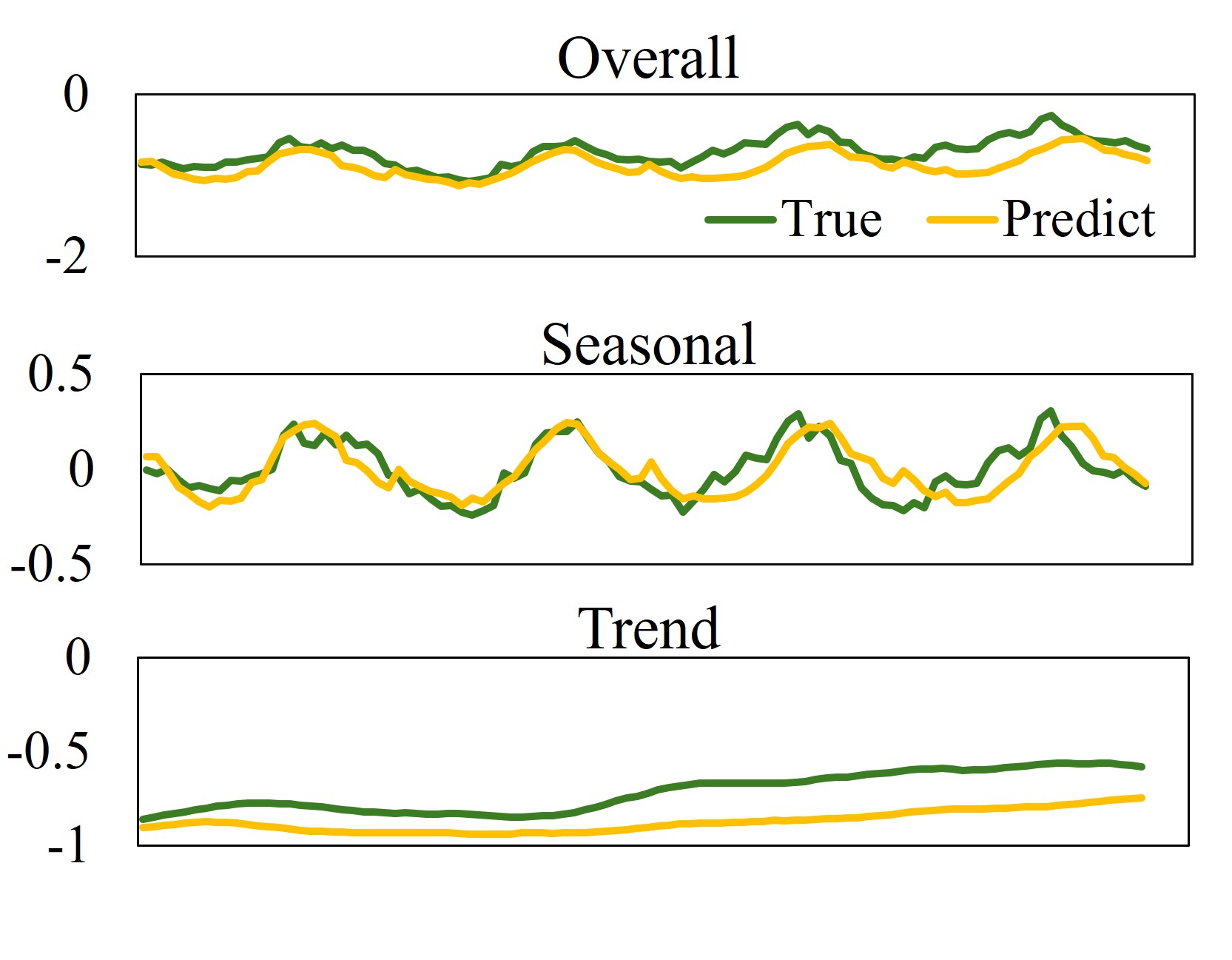}}
    \qquad
    \subfloat[Patchtst on ETTh2.]{\includegraphics[scale=0.32]{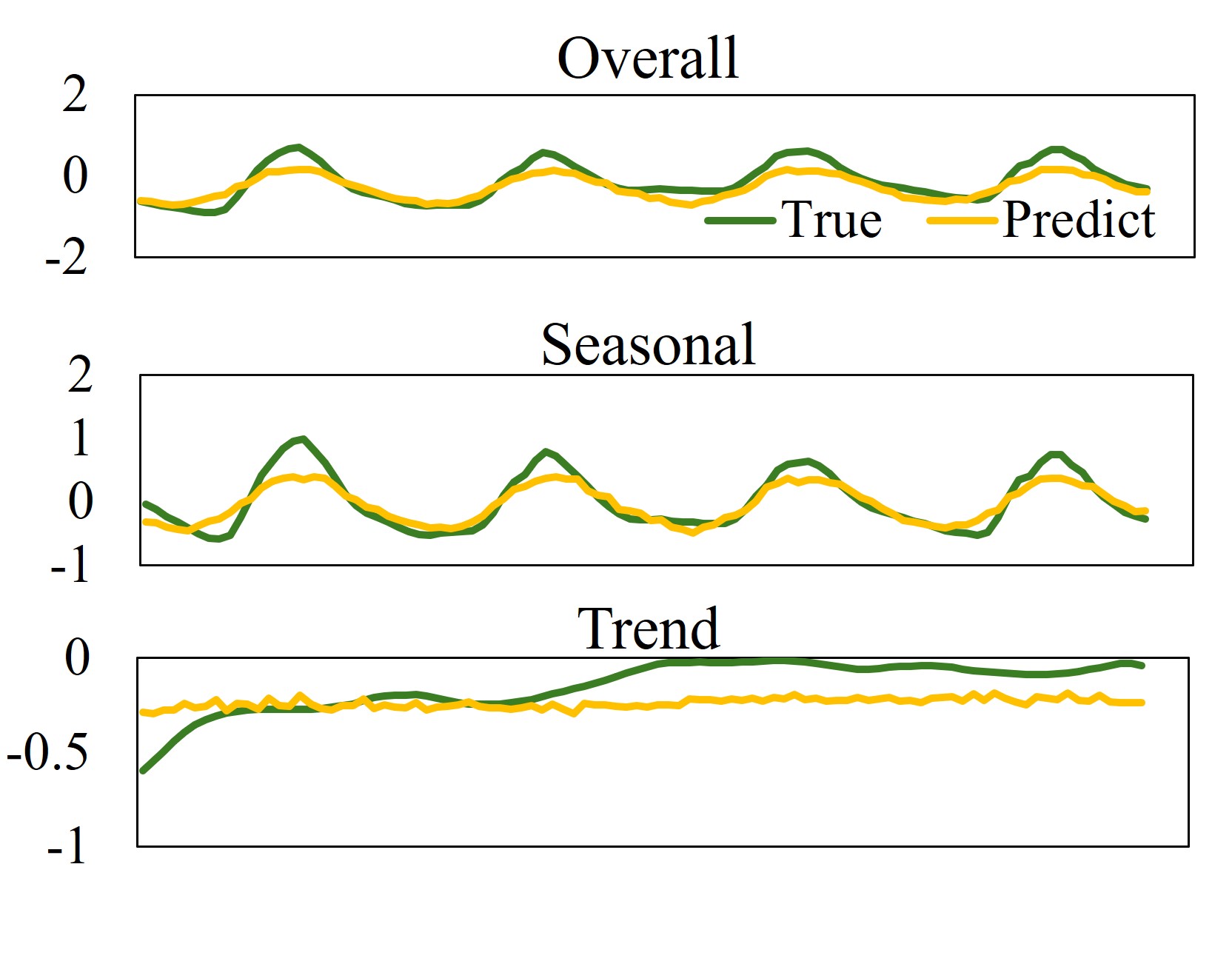}}
    \qquad
    \subfloat[Dlinear on ETTm1.]{\includegraphics[scale=0.32]{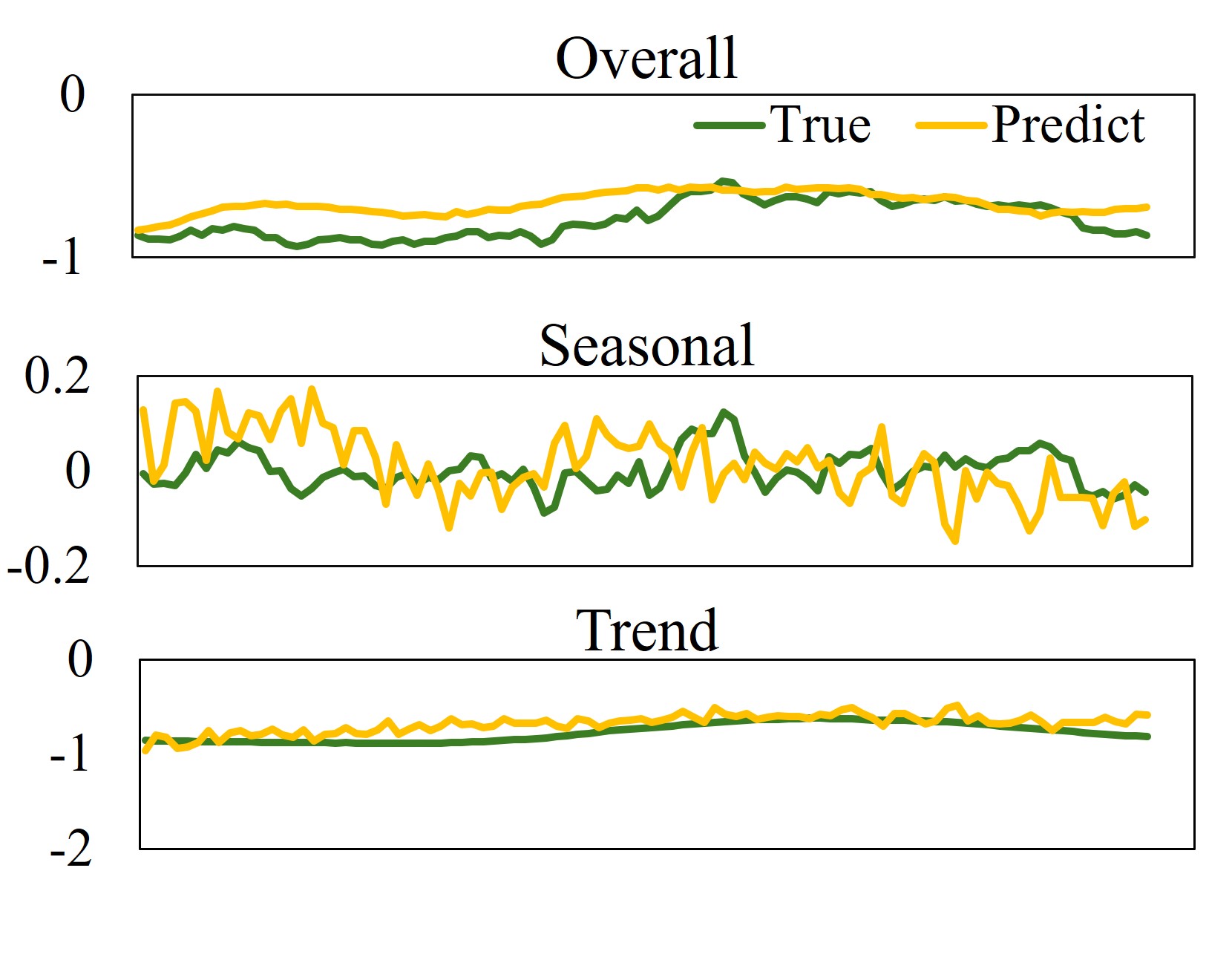}}
    \caption{The case study of time series forecasting. The results show the prediction-length-96 part (input length is 96) for different methods on different datasets. Each sub figure presents the single-variate (last variate) overall forecasting part and the forecasting part of the individual sub-series.}
    \label{fig:case}
\end{figure*}


\section{Method}
As revealed in the previous section, an overall loss indeed introduces bias when training deep learning methods on decomposed sub-series, potentially leading to significant errors, particularly in the Trend sub-series. To address this issue, we propose a hybrid loss framework in this section, which incorporates component-specific (sub-series) losses alongside the overall loss, and dynamically adjusts their weights to improve the overall and sub-series forecasting.

Specifically, we define the overall loss as $Loss_G$ (compute the MSE on the final results), the component loss as $Loss_C$, which is the sum of $Loss_S$ and $Loss_T$ for Seasonal and Trend sub-series (compute the MSE on the Seasonal sub-series and Trend sub-series results), respectively. We use a dual min-max problem to first balance the losses of $Loss_G$ and $Loss_C$, and then balance the losses of $Loss_S$ and $Loss_T$. This aims to maintain the overall forecasting performance while also focusing on and dynamically balancing the forecasting of sub-series with larger errors.


\subsection{Optimization Objective}
\label{Optimization Objective}
Drawing inspiration from distributionally robust optimization (DRO), our previous goal is to achieve optimal forecasting for the max loss part in our hybrid loss framework by adjusting the losses of $Loss_G$ and $Loss_C$, and the losses of $Loss_S$ and $Loss_T$ through dual min-max weighting. We define the first (min-max) optimization objective as follows:

\begin{equation}
    \min\limits_{\theta} \max\limits_{w_1 +w_2 =1, w_i \geq0} w_1 Loss_G +w_2 Loss_C,
    \label{opt1}
\end{equation}
where $\theta$ means the parameters of the deep learning method, $w_1$ and $w_2$ mean the weights for the overall loss $Loss_G$ and the component loss $Loss_C$ respectively, and the $Loss_C$ is associated with the second (min-max) optimization objective:

\begin{equation}
    Loss_C = \min\limits_{\theta} \max\limits_{\alpha + \beta =1, \alpha, \beta \geq0} \alpha Loss_S +\beta Loss_T,
    \label{opt2}
\end{equation}
where $\alpha$ and $\beta$ mean the weights for the Seasonal loss $Loss_S$ and the Trend loss $Loss_T$ respectively.

Therefore, the Equation (\ref{opt1}) means that when the component loss exceeds the overall loss, we need the model to prioritize the forecasting performance of sub-series rather than solely focusing on the final forecasting, and the Equation (\ref{opt2}) means that when optimizing for sub-series, we need the model to prioritize these with larger losses, as they are often the primary contributors to errors in the overall forecasting. We can combine these two optimization objectives as follows:

\begin{equation}
    \min\limits_{\theta} \max\limits_{w_1 +w_2 =1, w_i \geq0 \atop \alpha + \beta =1, \alpha, \beta \geq0} w_1 Loss_G +w_2 (\alpha Loss_S +\beta Loss_T).
    \label{opt}
\end{equation}







\subsection{Implementation}

To solve this optimization problem Equation (\ref{opt}), we also need to optimize the parameters $w_1$, $w_2$, $\alpha$ and $\beta$. Instead of applying the gradient descent method, we use estimation technique as the mirror descent method from DRO \citep{zhang2022towards} to update as follows:
\begin{equation}
    w_1 ^{cur}= \frac{w_1^{pre} \exp(\lambda_1 Loss_G)}{w_1^{pre}\exp(\lambda_1 Loss_G)+w_2^{pre} \exp(\lambda_1 Loss_C)}, 
\end{equation}

\begin{equation}
    w_2^{cur}= \frac{w_2^{pre} \exp(\lambda_1 Loss_C)}{w_1 \exp(\lambda_1 Loss_G)+w_2 ^{pre}\exp(\lambda_1 Loss_C)},
\end{equation}

\begin{equation}
    \alpha ^{cur}= \frac{\alpha ^{pre} \exp(\lambda_2 Loss_S)}{\alpha ^{pre} \exp(\lambda_2 Loss_S)+\beta^{pre} \exp(\lambda_2 Loss_T)} ,
\end{equation}

\begin{equation}
    \beta ^{cur}= \frac{\beta^{pre} \exp(\lambda_2 Loss_T)}{\alpha^{pre}  \exp(\lambda_2 Loss_S)+\beta^{pre} \exp(\lambda_2 Loss_T)} ,
\end{equation}
where $pre$ denotes the previous update step, $cur$ denotes the current update step. The $\lambda_i$ is a hyperparameter that balances the importance of the weighting term. Its value is often determined by the properties of the deep learning method being used. We initialize $w_1, w_2, \alpha, \beta =\frac{1}{2}$ in the initial iteration of our experiments. 

The optimization process then becomes: for each optimization step, we first compute the weights of the individual losses using the equations above, resulting in the combined loss
\begin{equation}
    Loss= w_1 Loss_G +w_2 (\alpha Loss_S +\beta Loss_T),
\end{equation}
which is then used to update the model parameters $\theta$ \footnote{The effectiveness and convergence of this optimization process are supported by prior work \citep{duchi2019variance}.}.

$$$$
\begin{table}[htbp!]
\centering
  \caption{Multivariate time series forecasting results on deep learning methods with/without hybrid loss framework. The ``Loss" indicates what kind of the loss does the methods use.}
  
\resizebox{0.92\textwidth}{!}{\begin{tabular}{cccccccc}
    \toprule
    \multirow{2}[4]{*}{Datasets} & Models & \multicolumn{2}{c}{Dlinear} & \multicolumn{2}{c}{FEDformer} & \multicolumn{2}{c}{Patchtst} \\
\cmidrule{2-8}          & Loss  & Original & Hybrid Loss & Original & Hybrid Loss & Original & Hybrid Loss \\
    \midrule
    \multirow{2}[2]{*}{ETTh1\newline{}} & MSE   & 0.4588  & \textbf{0.4579 } & 0.4394  & \textbf{0.4380 } & 0.4506  & \textbf{0.4502 } \\
          & MAE   & 0.4519  & \textbf{0.4511 } & 0.4581  & \textbf{0.4573 } & 0.4411  & \textbf{0.4402 } \\
    \midrule
    \multirow{2}[2]{*}{ETTh2\newline{}} & MSE   & 0.4981  & \textbf{0.4974 } & 0.4429  & \textbf{0.4417 } & 0.3658  & \textbf{0.3639 } \\
          & MAE   & 0.4792  & \textbf{0.4785 } & 0.4549  & \textbf{0.4539 } & 0.3945  & \textbf{0.3929 } \\
    \midrule
    \multirow{2}[2]{*}{ETTm1} & MSE   & 0.4061  & \textbf{0.4060 } & 0.4441  & \textbf{0.4424 } & 0.3838  & \textbf{0.3813 } \\
          & MAE   & \textbf{0.4102 } & \textbf{0.4102 } & 0.4543  & \textbf{0.4535 } & 0.3954  & \textbf{0.3943 } \\
    \midrule
    \multirow{2}[2]{*}{ETTm2} & MSE   & 0.3102  & \textbf{0.3100 } & 0.3031  & \textbf{0.3021 } & 0.2821  & \textbf{0.2790 } \\
          & MAE   & 0.3670  & \textbf{0.3667 } & 0.3493  & \textbf{0.3480 } & 0.3261  & \textbf{0.3247 } \\
    \midrule
    \multirow{2}[2]{*}{Electricity} & MSE   & 0.2095  & \textbf{0.2093 } & \textbf{0.2141 } & 0.2224  & \textbf{0.1951 } & 0.1955  \\
          & MAE   & 0.2956  & \textbf{0.2955 } & \textbf{0.3261 } & 0.3334  & \textbf{0.2794 } & 0.2796  \\
    \midrule
    \multirow{2}[2]{*}{Exchange} & MSE   & 0.3357  & \textbf{0.3307 } & \textbf{0.5017 } & 0.5201  & \textbf{0.3517 } & 0.3531  \\
          & MAE   & 0.3948  & \textbf{0.3947 } & \textbf{0.4908 } & 0.5025  & \textbf{0.3963 } & 0.3966  \\
    \midrule
    \multirow{2}[2]{*}{illness} & MSE   & 2.3465  & \textbf{2.3452 } & 2.7893  & \textbf{2.4759 } & 1.6318  & \textbf{1.5197 } \\
          & MAE   & \textbf{1.0883 } & 1.0892  & 1.1200  & \textbf{1.0974 } & 0.8616  & \textbf{0.8279 } \\
    \midrule
    \multirow{2}[2]{*}{Weather} & MSE   & 0.2670  & \textbf{0.2638 } & 0.3128  & \textbf{0.3112 } & \textbf{0.2598 } & 0.2605  \\
          & MAE   & 0.3174  & \textbf{0.3076 } & 0.3609  & \textbf{0.3589 } & 0.2816  & \textbf{0.2798 } \\
    \bottomrule
    \end{tabular}}
    \label{main table}
\end{table}

\section{Experiment}

In this section, we aim to validate the effectiveness of our proposed hybrid loss framework for both overall and sub-series forecasting performance across multiple datasets. We also conduct the ablation studies to analyze the contribution of each component of our loss framework.

\subsection{Experimental Setup}

\textbf{Datasets.} For the time series forecasting tasks, in addition to ETTh1, ETTh2, ETTm1, and ETTm2 datasets used in our preliminary experiments, we incorporate 4 more commonly used datasets: Electricity \citep{electricityloaddiagrams20112014_321}, Exchange-rate (Exchange) \citep{lai2018modeling}, National-illness (illness) \citep{zhou2021informer}, and Weather\footnote{https://www.bgc-jena.mpg.de/wetter/}, to demonstrate the broader applicability of our loss framework. These 4 datasets split into training, validation, and testing sets with the 3:1:1 ratio.

\textbf{Baseline.} Given that the models used in our preliminary experiments, DLinear \citep{zeng2023transformers}, Fedformer \citep{zhou2022fedformer}, and PatchTST \citep{nie2022time}, are already among the most prominent and effective, covering both MLP and transformer backbones, as well as point and patch embedding variants, we retain these models as baselines. Our method directly replaces the original loss function of these baselines with our proposed hybrid loss framework during training.\footnote{We also provide a comparison with a wider range of models in Appendix \ref{More Models}, demonstrating that models utilizing our hybrid loss framework still achieve state-of-the-art performance in a broader comparison.}

\textbf{Implementation details.}
In our experiments, except the nation-illness dataset, all the input lengths are 96, and prediction lengths are \{96, 192, 336, 720\}, respectively. For nation-illness dataset, the input length is 104 and prediction lengths are \{24, 36, 48, 60\}, respectively. To conserve space, the results presented in this section are averaged across all these prediction lengths.\footnote{More detailed results of each prediction length are provided in Appendix \ref{Results of each prediction length}.} Based on our validation set performance, we set $\lambda_1 = 0.9 $ and $\lambda_2 = 0.1$ for our hybrid loss framework across all models and datasets.
All experiments were conducted on a system with two NVIDIA V100 32G GPUs and an Intel(R) Xeon(R) CPU E5-2678 v3 @ 2.50GHz with 128GB of RAM. 

\textbf{Metrics.} We use the standard metrics of Mean Squared Error (MSE) and Mean Absolute Error (MAE) after the data normalization. A lower MSE and MAE indicates better performance.\footnote{More details of this section can be found in Appendix \ref{More Details}.}


\subsection{Main Results}

\textbf{Our hybrid loss framework effectively improves the final performance of existing methods across a wide range of datasets.} Table \ref{main table} presents the overall forecasting performance of these methods using both the original loss and our proposed hybrid loss framework. We observe improvements across most datasets. The magnitude of improvement is generally around 0.5-2\%, with a notable exception on the illness dataset where our method boosts the performance of FEDformer by nearly 10\% on MSE. This demonstrates that the dynamic focus on sub-series losses introduced by our hybrid loss framework is indeed effective and ultimately leads to improved overall performance.

We further investigate the reasons for the worse performance of FEDformer and PatchTST with our hybrid loss framework on the Electricity and Exchange-rate datasets. We find that the time series in these datasets lack readily discernible patterns and exhibit numerous abrupt changes. Consequently, incorporating sub-series losses reinforces the tendency to learn a smoother, low-frequency representation for each sub-series, which leads to less accurate forecasting in the final results.
\begin{table}[htbp!]
  \centering
  \caption{Multivariate time series forecasting overall and subseries results on deep learning methods with/without hybrid loss framework. The ``Global/Components" column indicates whether the reported results represent the overall forecasting performance or the performance on each individual decomposed sub-series. The ``Loss" column indicates what kind of the loss does the methods use.}
    \resizebox{0.95\textwidth}{!}{\begin{tabular}{ccccccccccc}
    \toprule
          &       &       & \multicolumn{2}{c}{ETTh1\newline{}} & \multicolumn{2}{c}{ETTh2\newline{}} & \multicolumn{2}{c}{ETTm1} & \multicolumn{2}{c}{ETTm2} \\
    \midrule
    Models & Loss  & Global/Componets & MSE   & MAE   & MSE   & MAE   & MSE   & MAE   & MSE   & MAE \\
    \midrule
    \multirow{6}[4]{*}{Dlinear} & \multirow{3}[2]{*}{Original} & Overall & 0.4588  & 0.4519  & 0.4981  & 0.4792  & 0.4061  & 0.4102  & 0.3102  & 0.3670  \\
          &       & Seasonal & 0.2965  & 0.3604  & 0.0888  & 0.2071  & 0.0969  & 0.2115  & 0.0486  & 0.1467  \\
          &       & Trend & 0.1716  & 0.3146  & 0.4144  & 0.4264  & 0.3192  & 0.3726  & 0.2661  & 0.3371  \\
\cmidrule{2-11}          & \multirow{3}[2]{*}{Hybrid} & Overall & 0.4579  & 0.4521  & 0.4974  & 0.4785  & 0.4060  & 0.4102  & 0.3100  & 0.3667  \\
          &       & Seasonal & 0.2923  & 0.3556  & 0.0819  & 0.1959  & 0.0961  & 0.2106  & 0.0435  & 0.1323  \\
          &       & Trend & 0.1686  & 0.3122  & 0.4038  & 0.4203  & 0.3189  & 0.3724  & 0.2611  & 0.3299  \\
    \midrule
    \multirow{6}[4]{*}{FEDformer} & \multirow{3}[2]{*}{Original} & Overall & 0.4394  & 0.4581  & 0.4429  & 0.4549  & 0.4441  & 0.4543  & 0.3031  & 0.3493  \\
          &       & Seasonal & 0.2793  & 0.3703  & 0.0866  & 0.2102  & 0.1010  & 0.2116  & 0.0446  & 0.1381  \\
          &       & Trend & 0.1678  & 0.3172  & 0.3551  & 0.4048  & 0.3249  & 0.3979  & 0.2595  & 0.3133  \\
\cmidrule{2-11}          & \multirow{3}[2]{*}{Hybrid} & Overall & 0.4380  & 0.4573  & 0.4417  & 0.4539  & 0.4424  & 0.4535  & 0.3021  & 0.3480  \\
          &       & Seasonal & 0.2786  & 0.3697  & 0.0826  & 0.2021  & 0.0944  & 0.2038  & 0.0415  & 0.1300  \\
          &       & Trend & 0.1649  & 0.3144  & 0.3409  & 0.4010  & 0.3187  & 0.3933  & 0.2570  & 0.3095  \\
    \midrule
    \multirow{6}[4]{*}{Patchtst} & \multirow{3}[2]{*}{Original} & Overall & 0.4506  & 0.4411  & 0.3658  & 0.3945  & 0.3838  & 0.3954  & 0.2821  & 0.3261  \\
          &       & Seasonal & 0.3031  & 0.3656  & 0.0835  & 0.1971  & 0.1319  & 0.2378  & 0.0490  & 0.1436  \\
          &       & Trend & 0.1566  & 0.2935  & 0.2710  & 0.3277  & 0.3198  & 0.3641  & 0.2328  & 0.2846  \\
\cmidrule{2-11}          & \multirow{3}[2]{*}{Hybrid} & Overall & 0.4502  & 0.4402  & 0.3699  & 0.3989  & 0.3813  & 0.3963  & 0.2790  & 0.3247  \\
          &       & Seasonal & 0.3008  & 0.3643  & 0.1116  & 0.1689  & 0.0880  & 0.1997  & 0.0451  & 0.1361  \\
          &       & Trend & 0.1521  & 0.2900  & 0.3020  & 0.3250  & 0.2746  & 0.3315  & 0.2304  & 0.2811  \\
    \bottomrule
    \end{tabular}}
  \label{tab:main2}%
\end{table}%

\textbf{Our hybrid loss framework significantly enhances the forecasting performance of individual sub-series, particularly the Trend sub-series.} Since our hybrid loss framework aims to improve overall performance by enhancing the forecasting of individual sub-series, we conduct additional experiments on the four datasets used in our preliminary experiments to compare the performance of our hybrid loss framework against the original loss, shown in Table \ref{tab:main2}. These results clearly demonstrate a significant reduction in the forecasting error of individual sub-series when using our hybrid loss framework. This improvement is particularly pronounced for the Trend sub-series, often exceeding a 2\% reduction in error. Given that most datasets exhibit greater forecasting deficiencies in the Trend component, and considering the importance of the Trend sub-series in representing the overall series trajectory, we believe our hybrid loss framework effectively addresses a common weakness in current decomposition-based deep learning methods.



\begin{figure*}[htb!]
    \centering


    \subfloat[FEDformer on ETTh1.]{\includegraphics[scale=0.32]{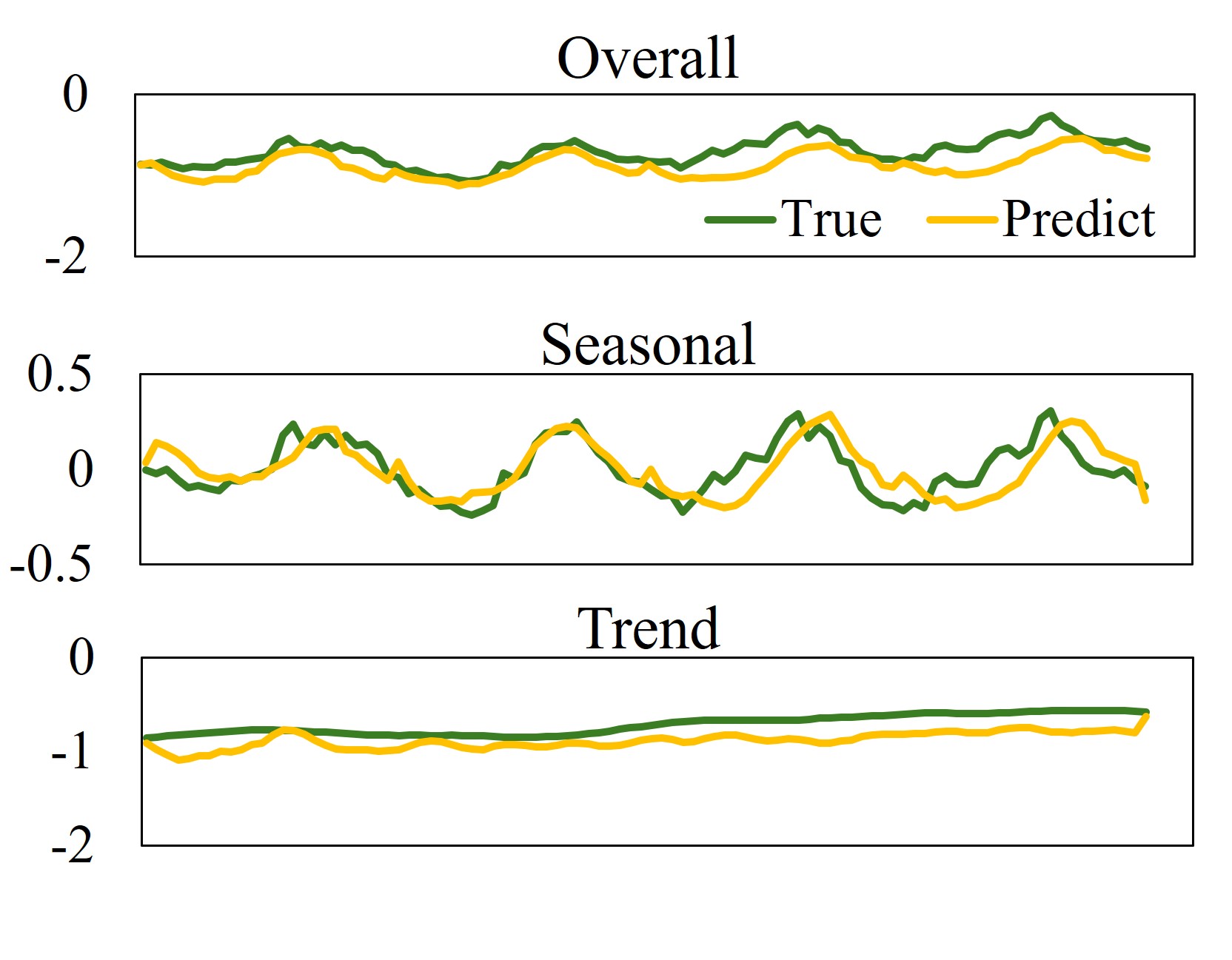}}
    \qquad
    \subfloat[Patchtst on ETTh2.]{\includegraphics[scale=0.32]{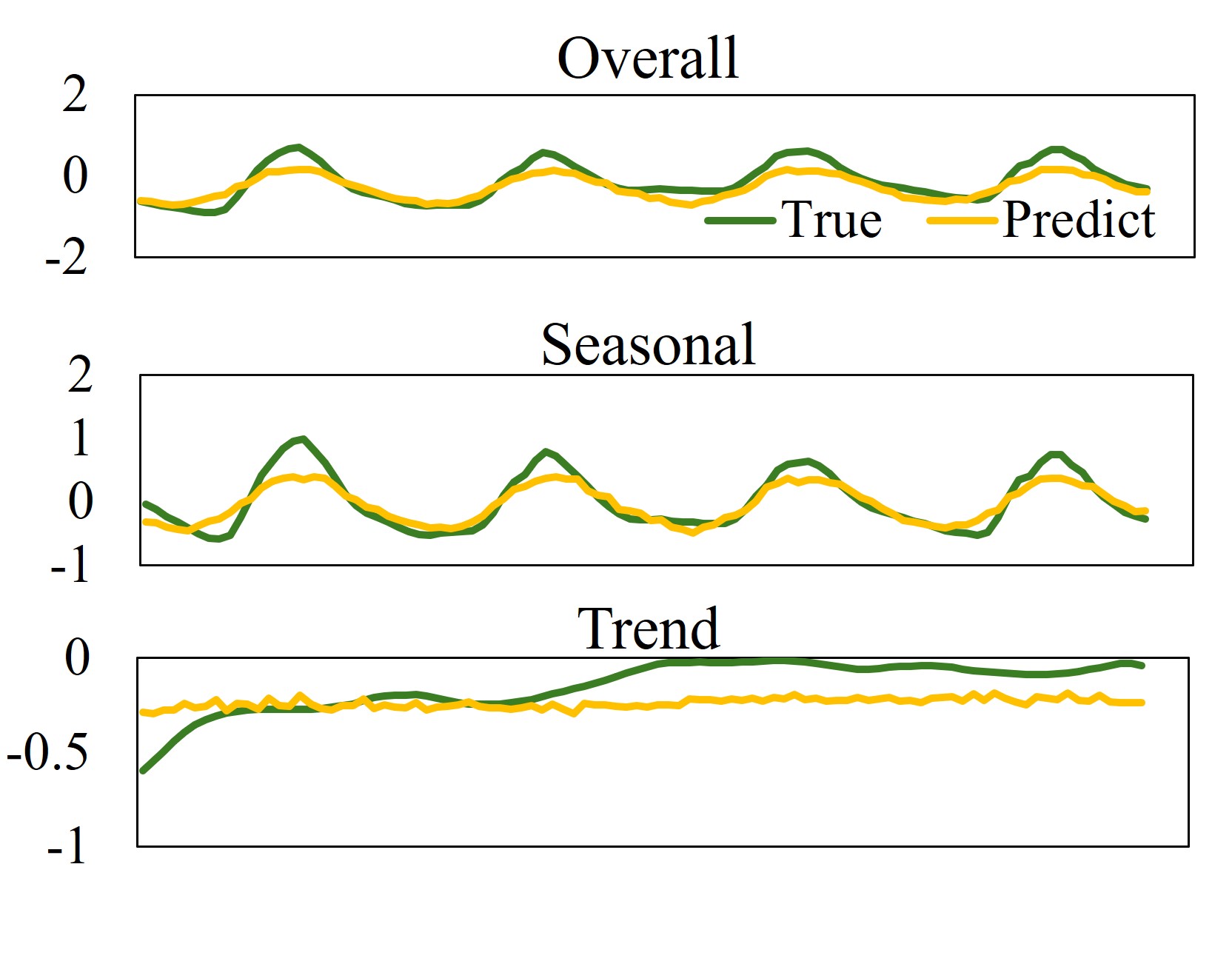}}
    \qquad
    \subfloat[Dlinear on ETTm1.]{\includegraphics[scale=0.32]{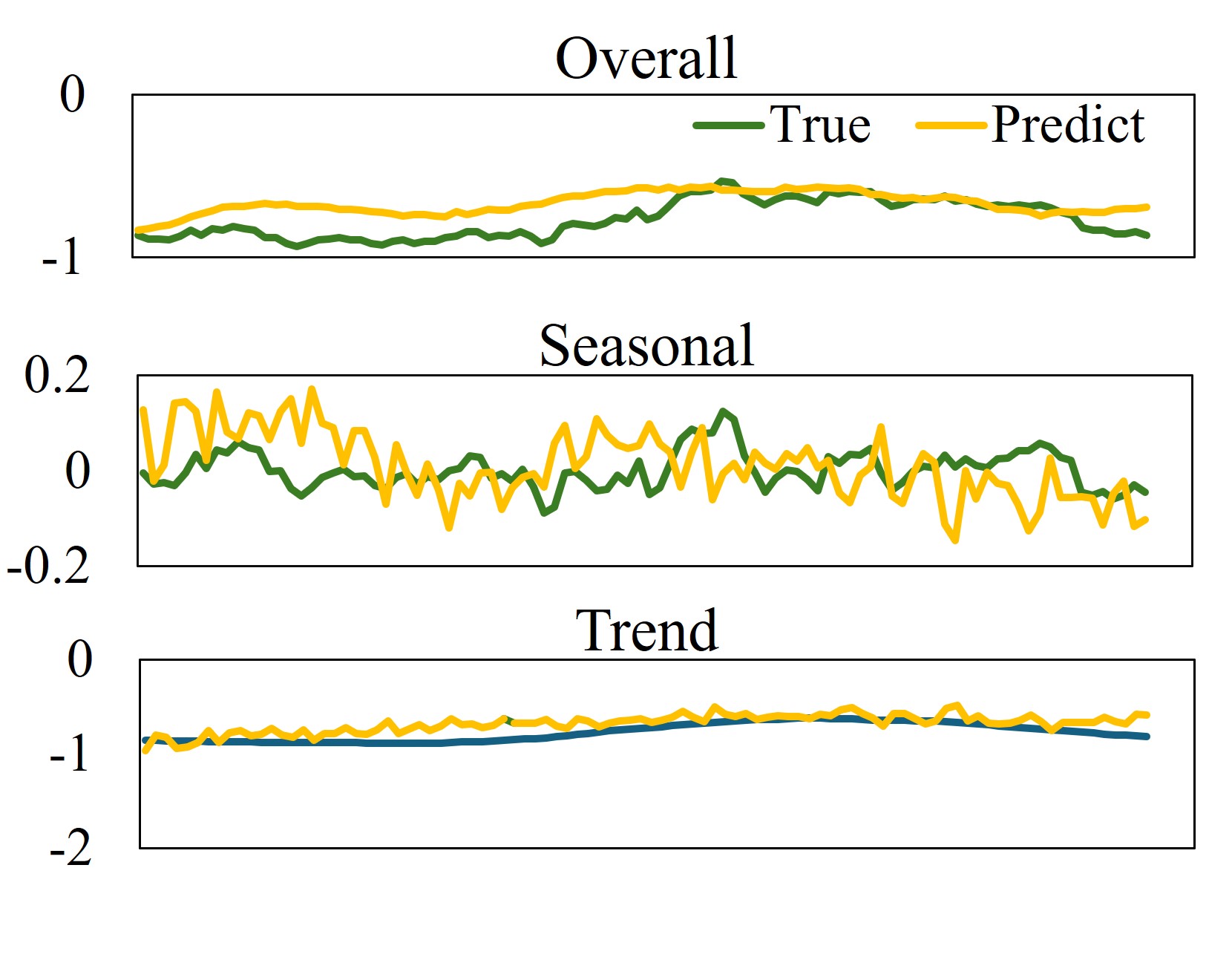}}
    \caption{The case study of time series forecasting results with our hybrid loss framework. The settings are the same as Figure \ref{fig:case}.}
    \label{fig:case-m}
\end{figure*}

We also perform the case study, using the same setup as in the preliminary experiments, to visually analyze the effects of our hybrid loss framework, as shown in Figure \ref{fig:case-m}.\footnote{More cases can be found in Appendix \ref{More Showcases}.} Comparing Figure \ref{fig:case} with Figure \ref{fig:case-m}, we observe a clear visual improvement in the forecasting of sub-series, particularly the Trend sub-series in Figure \ref{fig:case-m} (a) and (b). These now capture the upward trend, leading to better overall forecasting. DLinear on ETTm1 also shows a visually perceptible improvement in overall performance, despite some residual high-frequency errors in the Seasonal sub-series. Therefore, we believe learning the Trend sub-series may be a promising area for future research discovered by this work, and there is still room for further improvement even with the hybrid loss framework.

\subsection{Ablation Study}
To further explore our hybrid loss framework, we conduct two ablation studies: one to analyze the effectiveness of different components/variants, and another to assess the impact of varying initial hyperparameter settings.

\begin{table}[htbp!]
\centering
\caption{The ablation study of multivariate time series forecasting results on deep learning methods with our hybrid loss framework or its variants. The ``Loss" indicates what kind of the loss does the methods use.}
\resizebox{0.90\textwidth}{!}{\begin{tabular}{@{}cccccccccc@{}}
\toprule
\multicolumn{1}{c}{}       & \multicolumn{1}{c}{}     & \multicolumn{2}{c}{ETTh1} & \multicolumn{2}{c}{ETTh2} & \multicolumn{2}{c}{ETTm1} & \multicolumn{2}{c}{ETTm2} \\ \midrule
\multicolumn{1}{c}{Models} & \multicolumn{1}{c}{Loss} & MSE         & MAE         & MSE          & MAE        & MSE          & MAE        & MSE         & MAE         \\ \midrule
\multirow{3}{*}{Dlinear}    & Hybrid                   & \textbf{0.4579}      & \textbf{0.4521}      & \textbf{0.4974}       & \textbf{0.4785}     & 0.4060       & 0.4102     & \textbf{0.3100}      & \textbf{0.3667}      \\
                            & Component                 & 0.4593      & 0.4539      & 0.5603       & 0.5150     & \textbf{0.3978}       & \textbf{0.4101}     & 0.3239      & 0.3767      \\
                            & Fix weight               & 0.4596      & 0.4533      & 0.4993       & 0.4788     & 0.4079       & 0.4122     & 0.3164      & 0.3678      \\ \midrule
\multirow{3}{*}{FEDformer}  & Hybrid                   & \textbf{0.4380}      & \textbf{0.4573}      & \textbf{0.4417}       & \textbf{0.4539}     & \textbf{0.4424}       & \textbf{0.4535}     & \textbf{0.3021}      & \textbf{0.3480}      \\
                            & Component                 & 0.4879      & 0.4858      & 0.4574     & 0.4651     & 0.4778     & 0.4698    & 0.3251    & 0.3722    \\
                            & Fix weight               & 0.4414      & 0.4600      & 0.4456       & 0.4568     & 0.4670       & 0.4637     & 0.3055      & 0.3511      \\ \midrule
\multirow{3}{*}{Patchtst}   & Hybrid                   & \textbf{0.4502}      & \textbf{0.4402}      & 0.3699       & 0.3989     & \textbf{0.3813}       & \textbf{0.3963}     & \textbf{0.2790}      & 0.3247      \\
                            & Component                 & 0.4620      & 0.4498      & \textbf{0.3691}       & \textbf{0.3966}     & 0.3917       & 0.4001     & 0.2801      & \textbf{0.3240}      \\
                            & Fix weight               & 0.4586      & 0.4871      & 0.3742       & 0.4031     & 0.3900       & 0.3985     & 0.2792      & 0.3267      \\ \bottomrule
\end{tabular}}
\label{abla1}
\end{table}

For the first ablation study, we compare two variants of our hybrid loss framework in the datasets used in our preliminary experiments. The first variant, denoted as "Component", uses only the sub-series loss, corresponding to loss function Equation \ref{opt2}. The second variant, denoted as "Fixed Weight", uses fixed weights $w_1, w_2, \alpha, \beta$, all set to 0.5, during model training. The results of this ablation study are presented in Table \ref{abla1}.

\textbf{Using only the sub-series loss is insufficient, and dynamically updating the weights during training is crucial.} In Table \ref{abla1}, our hybrid loss framework achieves the best performance in most cases, often outperforming the "Component" variant (using only sub-series loss with min-max) by over 3\% and the "Fixed Weight" variant by 1-2\%. This demonstrates that solely focusing on the sub-series loss is insufficient; while the model may learn to predict sub-series well, the combined forecasting remains inaccurate. Furthermore, it highlights the dynamic nature of the balance between overall and sub-series losses during training, emphasizing that neither the overall loss nor any single sub-series consistently dominates the optimization process.

For the second ablation study, we compare the impact of different initial weights $w_1, w_2, \alpha, \beta$, to explore the influence of initial bias towards specific components of loss. As established in Section \ref{Optimization Objective}, $w_1 + w_2 = 1$ and $\alpha + \beta = 1$. Therefore, we test the following combinations: $w_1=0.1, \alpha=0.1$ or $0.9$; $w_1=0.5, \alpha=0.5$; and $w_1=0.9, \alpha=0.1$ or $0.9$. We used all datasets from the preliminary experiments and the DLinear and PatchTST models.\footnote{We also conduct this ablation study on FEDformer using the ETTh1 and ETTh2 datasets, presented in Appendix \ref{More Results of Ablation Study}.} The results of this ablation study are presented in Table \ref{abla2}.



\textbf{For most datasets, uniform initial weights (0.5) provide good performance, while excessive bias in the initial weights may lead to performance degradation.} As shown in Table \ref{abla2}, drastically altering the initial weights can still impact the final performance. The uniform initial weights (0.5) generally maintain stable performance and, in many cases, outperform initializations with 0.1 or 0.9 by approximately 1\%. This suggests that, in the absence of prior knowledge about the data, using a balanced set of initial weights (e.g., 0.5) for the our hybrid loss framework allows the model to learn and adjust these weights during training, leading to more reliable final performance compared to aggressively setting the initial weights.

\begin{table}[htbp!]
  \centering
  \caption{The ablation study of multivariate time series forecasting results on our hybrid loss framework with different initial weights. As $w_1 + w_2 = 1$ and $\alpha + \beta = 1$, we only specify the initial values of $w_1$ and $\alpha$ in the table.}
    \resizebox{0.90\textwidth}{!}{\begin{tabular}{ccccccccccc}
    \toprule
    \multicolumn{1}{c}{} &       &       & \multicolumn{2}{c}{ETTh1\newline{}} & \multicolumn{2}{c}{ETTh2\newline{}} & \multicolumn{2}{c}{ETTm1} & \multicolumn{2}{c}{ETTm2} \\
    \midrule
    Models & $w_1$  & $\alpha$ & MSE   & MAE   & MSE   & MAE   & MSE   & MAE   & MSE   & MAE \\
    \midrule
    \multirow{5}[6]{*}{Dlinear} & \multirow{2}[2]{*}{0.1} & 0.1   & 0.4596 & 0.4524 & 0.4939 & 0.4779 & 0.4091 & 0.4142 & 0.3064 & 0.3638 \\
          &       & 0.9   & 0.4610 & 0.4541 & 0.4978 & 0.4791 & 0.4047 & 0.4097 & 0.3091 & 0.3658 \\
\cmidrule{2-11}          & 0.5   & 0.5   & 0.4579 & 0.4511 & 0.4974 & 0.4785 & 0.4060 & 0.4102 & 0.3100 & 0.3667 \\
\cmidrule{2-11}          & \multirow{2}[2]{*}{0.9} & 0.1   & 0.4589 & 0.4519 & 0.4982 & 0.4793 & 0.4050 & 0.4103 & 0.3102 & 0.3670 \\
          &       & 0.9   & 0.4589 & 0.4519 & 0.4983 & 0.4793 & 0.4050 & 0.4103 & 0.3102 & 0.3670 \\
    \midrule
    \multirow{5}[6]{*}{Patchtst} & \multirow{2}[2]{*}{0.1} & 0.1   & 0.4554 & 0.4451 & 0.3651 & 0.3940 & 0.3836 & 0.3975 & 0.2802 & 0.3200 \\
          &       & 0.9   & 0.4518 & 0.4412 & 0.3650 & 0.3937 & 0.3828 & 0.3973 & 0.2800 & 0.3254 \\
\cmidrule{2-11}          & 0.5   & 0.5   & 0.4502 & 0.4402 & 0.3639 & 0.3929 & 0.3813 & 0.3943 & 0.2790 & 0.3247 \\
\cmidrule{2-11}          & \multirow{2}[2]{*}{0.9} & 0.1   & 0.4493 & 0.4393 & 0.3642 & 0.3935 & 0.3821 & 0.3960 & 0.2792 & 0.3247 \\
          &       & 0.9   & 0.4492 & 0.4391 & 0.3642 & 0.3934 & 0.3821 & 0.3935 & 0.2792 & 0.3248 \\
    \bottomrule
    \end{tabular}}
  \label{abla2}%
\end{table}%

\section{Related Works}

\textbf{The deep learning backbones in time series forecasting.}
Deep learning dominates time series forecasting in recent years. These methods leverage different powerful neural network architectures as backbones, adapting them to capture the characteristics of time series and learn effective predictive patterns from large datasets. For example, Prior to the rise of transformers, CNNs \citep{hewage2020temporal} and RNNs \citep{lai2018modeling} demonstrated the potential of deep learning to surpass traditional forecasting methods. Subsequently, transformers became the prevalent backbone, with models like Informer \citep{zhou2021informer}, Autoformer \citep{wu2021autoformer}, Fedformer \citep{zhou2022fedformer}, iTransformer \citep{liu2023itransformer}, and PatchTST \citep{nie2022time} specifically designed to exploit the sequential nature of time series. However, recent work suggests that simpler architectures, like MLP-based models such as DLinear \citep{zeng2023transformers}, TimeMixer \citep{wang2024timemixer}, and TimesNet \citep{wu2022timesnet}, can also achieve comparable or even superior performance. Furthermore, the impressive reasoning and generalization abilities of recent large language models (LLMs) \citep{jin2023time} have spurred exploration of their potential for zero-shot time series forecasting \citep{jin2023time,das2023decoder}. While these backbone architectures constitute the majority of time series forecasting research, many of them still employ time series decomposition techniques to better capture temporal dynamics by learning representations for individual sub-series. Furthermore, these models still rely on end-to-end overall loss functions\citep{jadon2024comprehensive}, leaving the relationship between the loss and the effectiveness of the learning of sub-series unexplored.

\textbf{The times series decomposition in time series forecasting.} Time series decomposition is a crucial component in many contemporary time series forecasting models, employed across various backbone architectures \citep{wu2021autoformer,zhou2022fedformer,nie2022time,zeng2023transformers}. Its core principle involves decomposing a raw time series into two or more sub-series, each representing specific characteristics of the original series. For example, the widely used sliding window approach \citep{faltermeier2010sliding} decomposes a time series into seasonal and trend components, capturing the local fluctuations and overall trajectory, respectively. Other models explore alternative decomposition methods based on mathematical principles. Fedformer \citep{zhou2022fedformer} builds upon the sliding window approach by further decomposing sub-series using Fourier transforms, focusing on dominant frequencies. TimeMixer \citep{wang2024timemixer} utilizes a multi-scale decomposition to capture information at different granularities. In this work, we specifically investigate how to enhance the learning of decomposed sub-series, particularly focusing on the commonly used sliding window decomposition method.

\section{Conclusion}
We explore the potential shortcomings of existing deep learning time series forecasting methods that incorporate time series decomposition. We find that the end-to-end overall loss employed by these methods may hinder the effective learning of decomposed sub-series, ultimately impacting the final performance.
Therefore, we propose a novel hybrid loss framework designed to address this balance between different sub-series and the overall series in time series forecasting. By employing a dual min-max loss framework, our approach dynamically emphasizes both the overall series and the sub-series that require enhanced learning. This avoids the bias that occurs when focusing solely on overall loss, which may lead to suboptimal model performance. 
Our framework achieves state-of-the-art performance across a wide range of datasets and experiments demonstrate that this loss framework can yield an average improvement of 0.5-2\% across existing time series models. 

\section*{Limitations and Future Work}
Despite the work presented in this study, from problem identification to solution development for decomposition-based deep learning methods in time series forecasting, our work has some limitations that we hope to address in future work.

First, although the investigated time series forecasting methods represent the current state-of-the-art, they all rely on sliding window decomposition. While alternative decomposition methods are less common, their performance under our loss framework may also need further investigation.

Second, due to computational constraints associated with averaging results across multiple prediction lengths for each datasets, the datasets used in our preliminary and ablation experiments could be expanded further to provide more comprehensive validation of our loss framework's effectiveness.

\section*{Reproducibility}

The code for our hybird loss framework is available in the Supplementary Material we submitted. It is designed as a plug-and-play module readily applicable to existing time series forecasting methods utilizing sliding window decomposition.


\bibliography{iclr2025_conference}

\begin{thebibliography}{33}
\providecommand{\natexlab}[1]{#1}
\providecommand{\url}[1]{\texttt{#1}}
\expandafter\ifx\csname urlstyle\endcsname\relax
  \providecommand{\doi}[1]{doi: #1}\else
  \providecommand{\doi}{doi: \begingroup \urlstyle{rm}\Url}\fi

\bibitem[Chen et~al.(2023)Chen, Li, Yoder, Arik, and Pfister]{chen2023tsmixer}
Si-An Chen, Chun-Liang Li, Nate Yoder, Sercan~O Arik, and Tomas Pfister.
\newblock Tsmixer: An all-mlp architecture for time series forecasting.
\newblock \emph{arXiv preprint arXiv:2303.06053}, 2023.

\bibitem[Cleveland et~al.(1990)Cleveland, Cleveland, McRae, Terpenning, et~al.]{cleveland1990stl}
Robert~B Cleveland, William~S Cleveland, Jean~E McRae, Irma Terpenning, et~al.
\newblock Stl: A seasonal-trend decomposition.
\newblock \emph{J. off. Stat}, 6\penalty0 (1):\penalty0 3--73, 1990.

\bibitem[Das et~al.(2023)Das, Kong, Sen, and Zhou]{das2023decoder}
Abhimanyu Das, Weihao Kong, Rajat Sen, and Yichen Zhou.
\newblock A decoder-only foundation model for time-series forecasting.
\newblock \emph{arXiv preprint arXiv:2310.10688}, 2023.

\bibitem[Duchi \& Namkoong(2019)Duchi and Namkoong]{duchi2019variance}
John Duchi and Hongseok Namkoong.
\newblock Variance-based regularization with convex objectives.
\newblock \emph{Journal of Machine Learning Research}, 20\penalty0 (68):\penalty0 1--55, 2019.

\bibitem[Faltermeier et~al.(2010)Faltermeier, Zeiler, Keck, Tom{\'e}, Brawanski, and Lang]{faltermeier2010sliding}
Rupert Faltermeier, Angela Zeiler, Ingo~R Keck, Ana~Maria Tom{\'e}, Alexander Brawanski, and Elmar~Wolfgang Lang.
\newblock Sliding empirical mode decomposition.
\newblock In \emph{The 2010 international joint conference on neural networks (IJCNN)}, pp.\  1--8. IEEE, 2010.

\bibitem[Godahewa et~al.(2021)Godahewa, Bergmeir, Webb, Hyndman, and Montero-Manso]{godahewa2021monash}
Rakshitha Godahewa, Christoph Bergmeir, Geoffrey~I Webb, Rob~J Hyndman, and Pablo Montero-Manso.
\newblock Monash time series forecasting archive.
\newblock \emph{arXiv preprint arXiv:2105.06643}, 2021.

\bibitem[Hewage et~al.(2020)Hewage, Behera, Trovati, Pereira, Ghahremani, Palmieri, and Liu]{hewage2020temporal}
Pradeep Hewage, Ardhendu Behera, Marcello Trovati, Ella Pereira, Morteza Ghahremani, Francesco Palmieri, and Yonghuai Liu.
\newblock Temporal convolutional neural (tcn) network for an effective weather forecasting using time-series data from the local weather station.
\newblock \emph{Soft Computing}, 24:\penalty0 16453--16482, 2020.

\bibitem[Jadon et~al.(2024)Jadon, Patil, and Jadon]{jadon2024comprehensive}
Aryan Jadon, Avinash Patil, and Shruti Jadon.
\newblock A comprehensive survey of regression-based loss functions for time series forecasting.
\newblock In \emph{International Conference on Data Management, Analytics \& Innovation}, pp.\  117--147. Springer, 2024.

\bibitem[Jin et~al.(2023)Jin, Wang, Ma, Chu, Zhang, Shi, Chen, Liang, Li, Pan, et~al.]{jin2023time}
Ming Jin, Shiyu Wang, Lintao Ma, Zhixuan Chu, James~Y Zhang, Xiaoming Shi, Pin-Yu Chen, Yuxuan Liang, Yuan-Fang Li, Shirui Pan, et~al.
\newblock Time-llm: Time series forecasting by reprogramming large language models.
\newblock \emph{arXiv preprint arXiv:2310.01728}, 2023.

\bibitem[Lai et~al.(2018)Lai, Chang, Yang, and Liu]{lai2018modeling}
Guokun Lai, Wei-Cheng Chang, Yiming Yang, and Hanxiao Liu.
\newblock Modeling long-and short-term temporal patterns with deep neural networks.
\newblock In \emph{The 41st international ACM SIGIR conference on research \& development in information retrieval}, pp.\  95--104, 2018.

\bibitem[Liu et~al.(2023)Liu, Hu, Zhang, Wu, Wang, Ma, and Long]{liu2023itransformer}
Yong Liu, Tengge Hu, Haoran Zhang, Haixu Wu, Shiyu Wang, Lintao Ma, and Mingsheng Long.
\newblock itransformer: Inverted transformers are effective for time series forecasting.
\newblock \emph{arXiv preprint arXiv:2310.06625}, 2023.

\bibitem[Namkoong \& Duchi(2016)Namkoong and Duchi]{namkoong2016stochastic}
Hongseok Namkoong and John~C Duchi.
\newblock Stochastic gradient methods for distributionally robust optimization with f-divergences.
\newblock \emph{Advances in neural information processing systems}, 29, 2016.

\bibitem[Nie et~al.(2022)Nie, Nguyen, Sinthong, and Kalagnanam]{nie2022time}
Yuqi Nie, Nam~H Nguyen, Phanwadee Sinthong, and Jayant Kalagnanam.
\newblock A time series is worth 64 words: Long-term forecasting with transformers.
\newblock \emph{arXiv preprint arXiv:2211.14730}, 2022.

\bibitem[OpenAI(2023)]{openai2023gpt}
R~OpenAI.
\newblock Gpt-4 technical report. arxiv 2303.08774.
\newblock \emph{View in Article}, 2\penalty0 (5), 2023.

\bibitem[Qian et~al.(2019{\natexlab{a}})Qian, Pei, Zareipour, and Chen]{QIAN2019939}
Zheng Qian, Yan Pei, Hamidreza Zareipour, and Niya Chen.
\newblock A review and discussion of decomposition-based hybrid models for wind energy forecasting applications.
\newblock \emph{Applied Energy}, 235:\penalty0 939--953, 2019{\natexlab{a}}.
\newblock ISSN 0306-2619.
\newblock \doi{https://doi.org/10.1016/j.apenergy.2018.10.080}.

\bibitem[Qian et~al.(2019{\natexlab{b}})Qian, Pei, Zareipour, and Chen]{qian2019review}
Zheng Qian, Yan Pei, Hamidreza Zareipour, and Niya Chen.
\newblock A review and discussion of decomposition-based hybrid models for wind energy forecasting applications.
\newblock \emph{Applied energy}, 235:\penalty0 939--953, 2019{\natexlab{b}}.

\bibitem[Trindade(2015)]{electricityloaddiagrams20112014_321}
Artur Trindade.
\newblock {ElectricityLoadDiagrams20112014}.
\newblock UCI Machine Learning Repository, 2015.
\newblock {DOI}: https://doi.org/10.24432/C58C86.

\bibitem[Vaswani(2017)]{vaswani2017attention}
Ashish Vaswani.
\newblock Attention is all you need.
\newblock \emph{arXiv preprint arXiv:1706.03762}, 2017.

\bibitem[Wang et~al.(2024)Wang, Wu, Shi, Hu, Luo, Ma, Zhang, and Zhou]{wang2024timemixer}
Shiyu Wang, Haixu Wu, Xiaoming Shi, Tengge Hu, Huakun Luo, Lintao Ma, James~Y Zhang, and Jun Zhou.
\newblock Timemixer: Decomposable multiscale mixing for time series forecasting.
\newblock \emph{arXiv preprint arXiv:2405.14616}, 2024.

\bibitem[Wiesemann et~al.(2014)Wiesemann, Kuhn, and Sim]{wiesemann2014distributionally}
Wolfram Wiesemann, Daniel Kuhn, and Melvyn Sim.
\newblock Distributionally robust convex optimization.
\newblock \emph{Operations research}, 62\penalty0 (6):\penalty0 1358--1376, 2014.

\bibitem[Woo et~al.(2022)Woo, Liu, Sahoo, Kumar, and Hoi]{woo2022etsformer}
Gerald Woo, Chenghao Liu, Doyen Sahoo, Akshat Kumar, and Steven Hoi.
\newblock Etsformer: Exponential smoothing transformers for time-series forecasting.
\newblock \emph{arXiv preprint arXiv:2202.01381}, 2022.

\bibitem[Wu et~al.(2021)Wu, Xu, Wang, and Long]{wu2021autoformer}
Haixu Wu, Jiehui Xu, Jianmin Wang, and Mingsheng Long.
\newblock Autoformer: Decomposition transformers with auto-correlation for long-term series forecasting.
\newblock \emph{Advances in neural information processing systems}, 34:\penalty0 22419--22430, 2021.

\bibitem[Wu et~al.(2022)Wu, Hu, Liu, Zhou, Wang, and Long]{wu2022timesnet}
Haixu Wu, Tengge Hu, Yong Liu, Hang Zhou, Jianmin Wang, and Mingsheng Long.
\newblock Timesnet: Temporal 2d-variation modeling for general time series analysis.
\newblock \emph{arXiv preprint arXiv:2210.02186}, 2022.

\bibitem[Wu et~al.(2023)Wu, Zhou, Long, and Wang]{wu2023interpretable}
Haixu Wu, Hang Zhou, Mingsheng Long, and Jianmin Wang.
\newblock Interpretable weather forecasting for worldwide stations with a unified deep model.
\newblock \emph{Nature Machine Intelligence}, 5\penalty0 (6):\penalty0 602--611, 2023.

\bibitem[Yin et~al.(2021)Yin, Wu, Wei, Shen, Qi, and Yin]{yin2021deep}
Xueyan Yin, Genze Wu, Jinze Wei, Yanming Shen, Heng Qi, and Baocai Yin.
\newblock Deep learning on traffic prediction: Methods, analysis, and future directions.
\newblock \emph{IEEE Transactions on Intelligent Transportation Systems}, 23\penalty0 (6):\penalty0 4927--4943, 2021.

\bibitem[Zeng et~al.(2023)Zeng, Chen, Zhang, and Xu]{zeng2023transformers}
Ailing Zeng, Muxi Chen, Lei Zhang, and Qiang Xu.
\newblock Are transformers effective for time series forecasting?
\newblock In \emph{Proceedings of the AAAI conference on artificial intelligence}, volume~37, pp.\  11121--11128, 2023.

\bibitem[Zhang et~al.(2022{\natexlab{a}})Zhang, Kuang, Liu, Chen, Lu, Wu, Wu, Xiao, et~al.]{zhang2022towards}
Fengda Zhang, Kun Kuang, Yuxuan Liu, Long Chen, Jiaxun Lu, Fei Wu, Chao Wu, Jun Xiao, et~al.
\newblock Towards multi-level fairness and robustness on federated learning.
\newblock In \emph{ICML 2022: Workshop on Spurious Correlations, Invariance and Stability}, 2022{\natexlab{a}}.

\bibitem[Zhang et~al.(2022{\natexlab{b}})Zhang, Zhang, Cao, Bian, Yi, Zheng, and Li]{zhang2022less}
Tianping Zhang, Yizhuo Zhang, Wei Cao, Jiang Bian, Xiaohan Yi, Shun Zheng, and Jian Li.
\newblock Less is more: Fast multivariate time series forecasting with light sampling-oriented mlp structures.
\newblock \emph{arXiv preprint arXiv:2207.01186}, 2022{\natexlab{b}}.

\bibitem[Zhang \& Yan(2023)Zhang and Yan]{zhang2023crossformer}
Yunhao Zhang and Junchi Yan.
\newblock Crossformer: Transformer utilizing cross-dimension dependency for multivariate time series forecasting.
\newblock In \emph{The eleventh international conference on learning representations}, 2023.

\bibitem[Zhou et~al.(2021{\natexlab{a}})Zhou, Zhang, Peng, Zhang, Li, Xiong, and Zhang]{haoyietal-informer-2021}
Haoyi Zhou, Shanghang Zhang, Jieqi Peng, Shuai Zhang, Jianxin Li, Hui Xiong, and Wancai Zhang.
\newblock Informer: Beyond efficient transformer for long sequence time-series forecasting.
\newblock In \emph{The Thirty-Fifth {AAAI} Conference on Artificial Intelligence, {AAAI} 2021, Virtual Conference}, volume~35, pp.\  11106--11115. {AAAI} Press, 2021{\natexlab{a}}.

\bibitem[Zhou et~al.(2021{\natexlab{b}})Zhou, Zhang, Peng, Zhang, Li, Xiong, and Zhang]{zhou2021informer}
Haoyi Zhou, Shanghang Zhang, Jieqi Peng, Shuai Zhang, Jianxin Li, Hui Xiong, and Wancai Zhang.
\newblock Informer: Beyond efficient transformer for long sequence time-series forecasting.
\newblock In \emph{Proceedings of the AAAI conference on artificial intelligence}, volume~35, pp.\  11106--11115, 2021{\natexlab{b}}.

\bibitem[Zhou et~al.(2022)Zhou, Ma, Wen, Wang, Sun, and Jin]{zhou2022fedformer}
Tian Zhou, Ziqing Ma, Qingsong Wen, Xue Wang, Liang Sun, and Rong Jin.
\newblock Fedformer: Frequency enhanced decomposed transformer for long-term series forecasting.
\newblock In \emph{International conference on machine learning}, pp.\  27268--27286. PMLR, 2022.

\bibitem[Zhou et~al.(2023)Zhou, Niu, Sun, Jin, et~al.]{zhou2023one}
Tian Zhou, Peisong Niu, Liang Sun, Rong Jin, et~al.
\newblock One fits all: Power general time series analysis by pretrained lm.
\newblock \emph{Advances in neural information processing systems}, 36:\penalty0 43322--43355, 2023.

\end{thebibliography}
\bibliographystyle{iclr2025_conference}

\newpage
\appendix
\section{More Details}
\label{More Details}
 We show more details of datasets, evaluation metrics, experiments in this section.

\textbf{Datasets details.} We evaluate the performance the methods on 8 commonly used datasets: ETTh1 \citep{haoyietal-informer-2021}, ETTh2 \citep{haoyietal-informer-2021}, ETTm1 \citep{zhou2021informer}, ETTm2 \citep{zhou2021informer}, Electricity \citep{electricityloaddiagrams20112014_321}, Exchange-rate (Exchange) \citep{lai2018modeling}, National-illness (illness) \citep{zhou2021informer}, and Weather\footnote{https://www.bgc-jena.mpg.de/wetter/}. Following the standard settings of the existing benchmarks \citep{zeng2023transformers,zhou2022fedformer,nie2022time}, except the nation-illness dataset, all the input lengths are 96, and prediction lengths are \{96, 192, 336, 720\}, respectively. For nation-illness dataset, the input length is 104 and prediction lengths are \{24, 36, 48, 60\}, respectively.
The first 4 datasets split into training, validation, and testing sets with the 7:1:2 ratio, and the last 4 datasets split into training, validation, and testing sets with the 3:1:1 ratio.
The detailed descriptions of these datasets in Table \ref{tab:data}.

\begin{table}[htbp!]
  \centering
  \caption{Dataset detailed descriptions. The dataset size is organized in (Train, Validation, Test). The ``Dim'' means the dimensions of the multivariate in the dataset.}
    \resizebox{0.95\textwidth}{!}{\begin{tabular}{c|c|c|c|c|c|c}
    \toprule
    Dataset & Dim   & Input Length & Prediction Length & Dataset Size & Frequency & Information \\
    \midrule
    \multirow{4}[8]{*}{ETTh1} & \multirow{4}[8]{*}{7} & \multirow{4}[8]{*}{96} & 96    & (8449, 2785, 2785) & \multirow{4}[8]{*}{Hourly} & \multirow{4}[8]{*}{Temperature} \\
\cmidrule{4-5}          &       &       & 192   & (8353, 2689, 2689) &       &  \\
\cmidrule{4-5}          &       &       & 336   & (8209, 2545, 2545) &       &  \\
\cmidrule{4-5}          &       &       & 720   & (7825, 2161, 2161) &       &  \\
    \midrule
    \multirow{4}[8]{*}{ETTh2} & \multirow{4}[8]{*}{7} & \multirow{4}[8]{*}{96} & 96    & (8449, 2785, 2785) & \multirow{4}[8]{*}{Hourly} & \multirow{4}[8]{*}{Temperature} \\
\cmidrule{4-5}          &       &       & 192   & (8353, 2689, 2689) &       &  \\
\cmidrule{4-5}          &       &       & 336   & (8209, 2545, 2545) &       &  \\
\cmidrule{4-5}          &       &       & 720   & (7825, 2161, 2161) &       &  \\
    \midrule
    \multirow{4}[8]{*}{ETTm1} & \multirow{4}[8]{*}{7} & \multirow{4}[8]{*}{96} & 96    & (34369, 11425, 11425) & \multirow{4}[8]{*}{15 mins} & \multirow{4}[8]{*}{Temperature} \\
\cmidrule{4-5}          &       &       & 192   & (34273, 11329, 11329) &       &  \\
\cmidrule{4-5}          &       &       & 336   & (34129, 11185, 11185) &       &  \\
\cmidrule{4-5}          &       &       & 720   & (33745, 10801, 10801) &       &  \\
    \midrule
    \multirow{4}[8]{*}{ETTm2} & \multirow{4}[8]{*}{7} & \multirow{4}[8]{*}{96} & 96    & (34369, 11425, 11425) & \multirow{4}[8]{*}{15 mins} & \multirow{4}[8]{*}{Temperature} \\
\cmidrule{4-5}          &       &       & 192   & (34273, 11329, 11329) &       &  \\
\cmidrule{4-5}          &       &       & 336   & (34129, 11185, 11185) &       &  \\
\cmidrule{4-5}          &       &       & 720   & (33745, 10801, 10801) &       &  \\
    \midrule
    \multirow{4}[8]{*}{Electricity} & \multirow{4}[8]{*}{321} & \multirow{4}[8]{*}{96} & 96    & (18221, 2537, 5165) & \multirow{4}[8]{*}{Hourly} & \multirow{4}[8]{*}{Electricity} \\
\cmidrule{4-5}          &       &       & 192   & (18125, 2441, 5069) &       &  \\
\cmidrule{4-5}          &       &       & 336   & (17981 , 2297, 4925) &       &  \\
\cmidrule{4-5}          &       &       & 720   & (17597, 1913, 4541) &       &  \\
    \midrule
    \multirow{4}[8]{*}{Exchange\_rate} & \multirow{4}[8]{*}{8} & \multirow{4}[8]{*}{96} & 96    & (5120, 665, 1422) & \multirow{4}[8]{*}{Day} & \multirow{4}[8]{*}{Exchange rates} \\
\cmidrule{4-5}          &       &       & 192   & (5024, 569, 1326) &       &  \\
\cmidrule{4-5}          &       &       & 336   & (4880, 425, 1182) &       &  \\
\cmidrule{4-5}          &       &       & 720   & (4496, 41, 798) &       &  \\
    \midrule
    \multirow{4}[8]{*}{illness} & \multirow{4}[8]{*}{7} & \multirow{4}[8]{*}{104} & 24    & (549, 74, 170) & \multirow{4}[8]{*}{Week} & \multirow{4}[8]{*}{National illness} \\
\cmidrule{4-5}          &       &       & 36    & (537, 62, 158) &       &  \\
\cmidrule{4-5}          &       &       & 48    & (525, 50, 146) &       &  \\
\cmidrule{4-5}          &       &       & 60    & (513, 38, 134) &       &  \\
    \midrule
    \multirow{4}[8]{*}{Weather} & \multirow{4}[8]{*}{21} & \multirow{4}[8]{*}{96} & 96    & (36696, 5175, 10444) & \multirow{4}[8]{*}{10min} & \multirow{4}[8]{*}{Weather} \\
\cmidrule{4-5}          &       &       & 192   & (36600, 5079, 10348) &       &  \\
\cmidrule{4-5}          &       &       & 336   & (36456, 4935, 10204) &       &  \\
\cmidrule{4-5}          &       &       & 720   & (36072, 4551, 9820) &       &  \\
    \bottomrule
    \end{tabular}}
  \label{tab:data}
\end{table}%

\textbf{Metric details.}
We utilize the mean square error (MSE) and mean absolute
error (MAE) for time series forecasting. The calculations of these metrics are:

\[ \textbf{MSE} =(\sum_{i=0}^{L}(\mathbf{Y_i} - \mathbf{\hat{Y_i}})^2)^{\frac{1}{2}},  \quad \textbf{MAE} = \sum_{i=1}^{L}|\mathbf{Y_i - \hat{Y_i}}|,  \]

where $\mathbf{Y, \hat{Y}} \in \mathbb{R}^{L\times C}$ are the  ground-truth and the forecasting results with $L$ time points and $C$ dimensions of multivariate, respectively. $\mathbf{Y_i}$ means the $i$th future time point.

\textbf{Experiment details.} Since we only modify the loss function, whose configuration is detailed in the main text, all other training parameters, including learning rate, batch size, epochs, and model-specific hyperparameters, are left at their default settings for each respective method. The original code for these methods is fully open-sourced in their respective original publications (we summarize the URL links of these models used in our paper in Table \ref{tab:modelsum}), allowing for straightforward reproduction.

\begin{table}[htbp!]
  \centering
  \caption{The URL links of the models we used in this paper.}
    \resizebox{0.98\textwidth}{!}{\begin{tabular}{c|c|c}
    \toprule
    \textbf{Model} & \textbf{Backbone} & \textbf{URL Link} \\
    \midrule
    TimeMixer & MLP & \url{https://github.com/kwuking/TimeMixer.git} \\
    \midrule
    TimesNet &  MLP &\url{https://github.com/thuml/TimesNet.git} \\
    \midrule
    Autoforemer  & Transformer & \url{https://github.com/thuml/Autoformer.git} \\
    \midrule
    Crossformer & Transformer &\url{https://github.com/Thinklab-SJTU/Crossformer.git} \\
    \midrule
    iTransformer & Transformer &\url{https://github.com/thuml/iTransformer.git} \\
    \midrule
    GPT2  & LLM (Transformer) &\url{https://github.com/DAMO-DI-ML/NeurIPS2023-One-Fits-All.git} \\
    \midrule
    TimesFM & LLM (Transformer) &\url{https://github.com/google-research/timesfm.git} \\
    \midrule
    Dlinear  & MLP &\url{https://github.com/cure-lab/LTSF-Linear.git} \\
    \midrule
    FEDformer & Transformer &\url{https://github.com/MAZiqing/FEDformer.git} \\
    \midrule
    PatchTST & Transformer &\url{https://github.com/yuqinie98/PatchTST.git} \\
    \bottomrule
    \end{tabular}}
  \label{tab:modelsum}%
\end{table}%

\section{More Results of Main Experients}

\subsection{More Models}
\label{More Models}
In addition to the 3 models compared in the main text, we include 7 more models for a broader comparison. These include two MLP-based models: TimeMixer \citep{wang2024timemixer} and TimesNet \citep{wu2022timesnet}; three Transformer-based models: Autoformer \citep{wu2021autoformer}, Crossformer \citep{zhang2023crossformer}, and iTransformer \citep{liu2023itransformer}; and two recent LLM-based models: GPT2 \citep{zhou2023one} and TimesFM \citep{das2023decoder}. The results are presented in Table \ref{tab:More Models}.

\begin{table}[htbp!]
\caption{Multivariate time series forecasting results on more deep learning methods with/without hybrid loss framework.}
\resizebox{\textwidth}{!}{\begin{tabular}{ccccccccccccccc}
    \toprule
     Models     & Metrics & TimeMixer & TimesNet & Autoforemer & Crossformer & iTransformer & GPT2  & TimesFM & Dlinear & \multicolumn{1}{p{4.5em}}{Dlinear\newline{}(Hybrid Loss)} & FEDformer & \multicolumn{1}{p{4.5em}}{FEDformer\newline{}(Hybrid Loss)} & PatchTST & \multicolumn{1}{p{4.5em}}{PatchTST\newline{}(Hybrid Loss)} \\
    \midrule
    \multirow{2}[2]{*}{ETTh1\newline{}} & MSE   & 0.4512 & 0.4609 & 0.4738 & 0.5987 & 0.4570 & 0.4681 & 0.5406 & 0.4588 & 0.4579 & 0.4394 & \textbf{0.4380} & 0.4506 & 0.4502 \\
          & MAE   & 0.4405 & 0.4551 & 0.4733 & 0.5586 & 0.4492 & 0.4558 & 0.4446 & 0.4519 & 0.4511 & 0.4581 & 0.4573 & 0.4411 & \textbf{0.4402} \\
    \midrule
    \multirow{2}[2]{*}{ETTh2\newline{}} & MSE   & 0.3849 & 0.4074 & 0.4258 & 0.5662 & 0.3837 & 0.3792 & \textbf{0.3127} & 0.4981 & 0.4974 & 0.4429 & 0.4417 & 0.3658 & 0.3639 \\
          & MAE   & 0.4061 & 0.4211 & 0.4447 & 0.5451 & 0.4069 & 0.4054 & \textbf{0.3748} & 0.4792 & 0.4785 & 0.4549 & 0.4539 & 0.3945 & 0.3929 \\
    \midrule
    \multirow{2}[2]{*}{ETTm1} & MSE   & 0.3908 & 0.4101 & 0.5502 & 0.5065 & 0.4076 & 0.3875 & 0.5240 & 0.4061 & 0.4060 & 0.4441 & 0.4424 & 0.3838 & \textbf{0.3813} \\
          & MAE   & 0.4023 & 0.4177 & 0.5024 & 0.5030 & 0.4118 & 0.4020 & 0.4577 & 0.4102 & 0.4102 & 0.4543 & 0.4535 & 0.3954 & \textbf{0.3943} \\
    \midrule
    \multirow{2}[2]{*}{ETTm2} & MSE   & \textbf{0.2767} & 0.2950 & 0.3251 & 1.5484 & 0.2922 & 0.2852 & 0.3390 & 0.3102 & 0.3100 & 0.3031 & 0.3021 & 0.2821 & 0.2790 \\
          & MAE   & \textbf{0.3232} & 0.3317 & 0.3637 & 0.7716 & 0.3358 & 0.3287 & 0.3586 & 0.3670 & 0.3667 & 0.3493 & 0.3480 & 0.3261 & 0.3247 \\
    \midrule
    \multirow{2}[2]{*}{Electricity} & MSE   & 0.1818 & 0.1941 & 0.2370 & 0.3065 & 0.1756 & \textbf{0.1626} & 0.1860 & 0.2095 & 0.2093 & 0.2141 & 0.2224 & 0.1951 & 0.1955 \\
          & MAE   & 0.2722 & 0.2956 & 0.3436 & 0.3583 & 0.2666 & \textbf{0.2558} & 0.2667 & 0.2956 & 0.2955 & 0.3261 & 0.3334 & 0.2794 & 0.2796 \\
    \midrule
    \multirow{2}[2]{*}{Exchange} & MSE   & 0.4356 & 0.4093 & 0.4901 & 0.9711 & 0.3642 & 0.3624 & \textbf{0.2313} & 0.3357 & 0.3307 & 0.5017 & 0.5201 & 0.3517 & 0.3531 \\
          & MAE   & 0.4298 & 0.4403 & 0.4929 & 0.7315 & 0.4069 & 0.4065 & \textbf{0.3328} & 0.3948 & 0.3947 & 0.4908 & 0.5025 & 0.3963 & 0.3966 \\
    \midrule
    \multirow{2}[2]{*}{illness} & MSE   & 1.7500 & 2.2410 & 3.0330 & 3.7904 & 2.1360 & 1.9338 & 2.8652 & 2.3465 & 2.3452 & 2.7893 & 2.4759 & 1.6318 & \textbf{1.5197} \\
          & MAE   & 0.8706 & 0.9234 & 1.2053 & 1.2825 & 1.0075 & 0.9016 & 1.1173 & 1.0883 & 1.0892 & 1.1200 & 1.0974 & 0.8616 & \textbf{0.8279} \\
    \midrule
    \multirow{2}[2]{*}{Weather} & MSE   & \textbf{0.2459} & 0.2588 & 0.3366 & 0.2638 & 0.2598 & 0.2548 & 0.2750 & 0.2670 & 0.2638 & 0.3128 & 0.3112 & 0.2598 & 0.2605 \\
          & MAE   & \textbf{0.2750} & 0.2857 & 0.3825 & 0.3229 & 0.2805 & 0.2780 & 0.2788 & 0.3174 & 0.3076 & 0.3609 & 0.3589 & 0.2816 & 0.2798 \\
    \bottomrule
    \end{tabular}%
}  \label{tab:More Models}
\end{table}

\textbf{Even with the increasing prevalence of LLM-based time series forecasting methods, our hybrid loss framework still enables existing models to achieve state-of-the-art performance in most cases.} The results in Table \ref{tab:More Models} demonstrate that, across 8 datasets, methods using our hybrid loss framework achieve state-of-the-art performance on 3 datasets, matching the number achieved by LLM-based methods and tying for the overall lead. This highlights the significant improvements provided by our hybrid loss framework for existing non-LLM methods and further suggests that there is still room for improvement in these methods.

\subsection{Results of each Prediction Length}
\label{Results of each prediction length}
Here, we present the results for each prediction length in our main experiment.

\begin{table}[htbp!]
\centering
  \caption{Multivariate time series forecasting results on deep learning methods with/without hybrid loss framework (prediction
length is 96). }
  
\resizebox{0.92\textwidth}{!}{\begin{tabular}{cccccccc}
    \toprule
    \multirow{2}[4]{*}{Datasets} & Models & \multicolumn{2}{c}{Dlinear} & \multicolumn{2}{c}{FEDformer} & \multicolumn{2}{c}{Patchtst} \\
\cmidrule{2-8}          & Loss  & Original & Hybrid Loss & Original & Hybrid Loss & Original & Hybrid Loss \\
    \midrule
    \multirow{2}[2]{*}{ETTh1\newline{}} & MSE   & 0.3829  & \textbf{0.3779} & 0.3771  & \textbf{0.3770} & 0.3935 & \textbf{0.3924} \\
          & MAE   & \textbf{0.3959}  & 0.3960 & 0.4185  & \textbf{0.4184} & 0.4080  & \textbf{0.4061} \\
    \midrule
    \multirow{2}[2]{*}{ETTh2\newline{}} & MSE   & 0.3290  & \textbf{0.3279} & 0.3508 & \textbf{0.3481} & 0.2938  & \textbf{0.2927 } \\
          & MAE   & 0.3804  &\textbf{ 0.3795} & 0.3918 & \textbf{0.3902} & 0.3427  & \textbf{0.3415 } \\
    \midrule
    \multirow{2}[2]{*}{ETTm1} & MSE   & 0.3458  & \textbf{0.3457 } & 0.3669  & \textbf{0.3628 } & 0.3211  & \textbf{0.3183} \\
          & MAE   & \textbf{0.3737 } & \textbf{0.3737 } & 0.4122  & \textbf{0.4097 } & 0.3596 & \textbf{0.3572 } \\
    \midrule
    \multirow{2}[2]{*}{ETTm2} & MSE   &\textbf{ 0.1869} & \textbf{0.1869} & 0.1918  & \textbf{0.1908 } & 0.1776 & \textbf{0.1758 } \\
          & MAE   & 0.2811  & \textbf{0.2810 } & 0.2812  & \textbf{0.2801 } &0.2599  & \textbf{0.2586 } \\
    \midrule
    \multirow{2}[2]{*}{Electricity} & MSE   & 0.1946  & \textbf{0.1944 } & \textbf{0.1884 } & 0.1950  & 0.1718  & \textbf{0.1664}  \\
          & MAE   & 0.2774  & \textbf{0.2773 } & \textbf{0.3036 } & 0.3095  & 0.2573  & \textbf{0.2332}  \\
    \midrule
    \multirow{2}[2]{*}{Exchange} & MSE   & 0.0782  & \textbf{0.0779 } & \textbf{0.1447 } & 0.16653  & 0.0806  & \textbf{0.0805}  \\
          & MAE   & 0.1985  & \textbf{0.1977 } & \textbf{0.2736 } & 0.2942  & 0.1973  & \textbf{0.1965}  \\
    \midrule
    \multirow{2}[2]{*}{illness} & MSE   & 2.2795  & \textbf{2.2794 } & 3.2211  & \textbf{2.7505 } & 1.7609  & \textbf{1.4334 } \\
          & MAE   & \textbf{1.0601 } & 1.0622  & 1.2420  & \textbf{1.1599 } & 0.9018  & \textbf{0.8020 } \\
    \midrule
    \multirow{2}[2]{*}{Weather} & MSE   & 0.1969  & \textbf{0.1968 } & \textbf{0.2231}  & 0.2232  & 0.1816  & \textbf{0.1769}  \\
          & MAE   & 0.2551  & \textbf{0.2550 } & \textbf{0.3051}  & 0.3061  & 0.2219  & \textbf{0.2157 } \\
    \bottomrule
    \end{tabular}}
    \label{main table1}
\end{table}

\begin{table}[htbp!]
\centering
  \caption{Multivariate time series forecasting results on deep learning methods with/without hybrid loss framework (prediction
length is 192). }
  
\resizebox{0.92\textwidth}{!}{\begin{tabular}{cccccccc}
    \toprule
    \multirow{2}[4]{*}{Datasets} & Models & \multicolumn{2}{c}{Dlinear} & \multicolumn{2}{c}{FEDformer} & \multicolumn{2}{c}{Patchtst} \\
\cmidrule{2-8}          & Loss  & Original & Hybrid Loss & Original & Hybrid Loss & Original & Hybrid Loss \\
    \midrule
    \multirow{2}[2]{*}{ETTh1\newline{}} & MSE   & \textbf{0.4327} & \textbf{0.4327} & 0.4200  &\textbf{ 0.4198} & \textbf{0.4453} & 0.4464 \\
          & MAE   & \textbf{0.4258}  & \textbf{0.4258} & 0.4441  & \textbf{0.4439} & 0.4342  & \textbf{0.4338} \\
    \midrule
    \multirow{2}[2]{*}{ETTh2\newline{}} & MSE   & \textbf{0.4313} & 0.4333 & 0.4420& \textbf{0.4407} & 0.3769  & \textbf{0.3744 } \\
          & MAE   & \textbf{0.4432}  & 0.4446& 0.4498  & \textbf{0.4482} & 0.3930  & \textbf{0.3913 } \\
    \midrule
    \multirow{2}[2]{*}{ETTm1} & MSE   & 0.3826  & \textbf{0.3825 } & 0.4360  & \textbf{0.4345} & 0.3652  & \textbf{0.3625} \\
          & MAE   & 0.3929  & \textbf{0.3928} & 0.4465  & \textbf{0.4453} & \textbf{0.3820}& 0.3828  \\
    \midrule
    \multirow{2}[2]{*}{ETTm2} & MSE   & 0.2720 & \textbf{0.2712} & 0.2637  & \textbf{0.2636} & 0.2487 & \textbf{0.2408 } \\
          & MAE   & 0.3486  & \textbf{0.3477 } & 0.3255  & \textbf{0.3252 } & 0.3064  & \textbf{0.3023 } \\
    \midrule
    \multirow{2}[2]{*}{Electricity} & MSE   & \textbf{0.1939}  & \textbf{0.1939 } & \textbf{0.1964 } & 0.2023  & \textbf{0.1789 } & 0.1824  \\
          & MAE   & \textbf{0.2804}  & \textbf{0.2804 } & \textbf{0.3109 } & 0.3156  & \textbf{0.2647 } & 0.2685  \\
    \midrule
    \multirow{2}[2]{*}{Exchange} & MSE   & \textbf{0.1559}  & 0.1562  & \textbf{0.2648 } & 0.2706  & 0.1710  & \textbf{0.1704}  \\
          & MAE   & \textbf{0.2921}  & 0.2926  & \textbf{0.3745 } & 0.3812  & 0.2931  & \textbf{0.2920}  \\
    \midrule
    \multirow{2}[2]{*}{illness} & MSE   & 2.2350  & \textbf{2.2323 }& 2.5884  & \textbf{2.3293}  & \textbf{1.4001}  & \textbf1.5344  \\
          & MAE   & \textbf{1.0580 } & 1.0586  & 1.1204  & \textbf{1.0973 } & 0.8616  & \textbf{0.8279 } \\
    \midrule
    \multirow{2}[2]{*}{Weather} & MSE   & 0.2392  & \textbf{0.2265 } & 0.2847  & \textbf{0.2782 } & \textbf{0.2271 } & 0.2275  \\
          & MAE   & 0.2971  & \textbf{0.2582 } & 0.3547  & \textbf{0.3454 } & 0.2601  & \textbf{0.2582 } \\
    \bottomrule
    \end{tabular}}
    \label{main table2}
\end{table}

\begin{table}[htbp!]
\centering
  \caption{Multivariate time series forecasting results on deep learning methods with/without hybrid loss framework (prediction
length is 336). }
  
\resizebox{0.92\textwidth}{!}{\begin{tabular}{cccccccc}
    \toprule
    \multirow{2}[4]{*}{Datasets} & Models & \multicolumn{2}{c}{Dlinear} & \multicolumn{2}{c}{FEDformer} & \multicolumn{2}{c}{Patchtst} \\
\cmidrule{2-8}          & Loss  & Original & Hybrid Loss & Original & Hybrid Loss & Original & Hybrid Loss \\
    \midrule
    \multirow{2}[2]{*}{ETTh1\newline{}} & MSE   & 0.4913
  & \textbf{0.4912} & 0.4581  & \textbf{0.4577} & \textbf{0.4838} & 0.4843 \\
          & MAE   & 0.4673 & \textbf{0.4671} & 0.4664  & \textbf{0.4659} & 0.4515  & \textbf{0.4511} \\
    \midrule
    \multirow{2}[2]{*}{ETTh2\newline{}} & MSE   & \textbf{0.4586}  & 0.4604 & 0.4985 & \textbf{0.4982} & 0.3806  & \textbf{0.3800 } \\
          & MAE   & \textbf{0.4618}  & 0.4633 & 0.4905 & \textbf{0.4902} & \textbf{0.4089}  & 0.4091  \\
    \midrule
    \multirow{2}[2]{*}{ETTm1} & MSE   & \textbf{0.4165}  & \textbf{0.4165 } & 0.4666  & \textbf{0.4659 } & 0.3933  & \textbf{0.3910} \\
          & MAE   & \textbf{0.4175 } & \textbf{0.4175 } & \textbf{0.4699}  & 0.4702  & \textbf{0.4039} & 0.4052  \\
    \midrule
    \multirow{2}[2]{*}{ETTm2} & MSE   & 0.3434 & \textbf{0.3433} & 0.3306  & \textbf{0.3267 } & 0.3033 & \textbf{0.3027 } \\
          & MAE   & \textbf{0.3945}  & \textbf{0.3945 } & 0.3674  & \textbf{0.3637 } &\textbf{0.3411}  & \textbf0.3423  \\
    \midrule
    \multirow{2}[2]{*}{Electricity} & MSE   &\textbf{ 0.2069}  & \textbf{0.2069 } & \textbf{0.2076 } & 0.2289  & \textbf{0.1946 } & 0.1975  \\
          & MAE   & 0.2963  & \textbf{0.2963 } & \textbf{0.3231 } & 0.3420  & \textbf{0.2811 } & 0.2893  \\
    \midrule
    \multirow{2}[2]{*}{Exchange} & MSE   & 0.3269  & \textbf{0.3071 } & \textbf{0.4437 } & 0.4740  & \textbf{0.3188 } & 0.3202  \\
          & MAE   & \textbf{0.4192}  & 0.4194  & \textbf{0.4923 } & 0.5049  & \textbf{0.4070 } & 0.4078  \\
    \midrule
    \multirow{2}[2]{*}{illness} & MSE   & 2.2983  & \textbf{2.2925 } & 2.5682  & \textbf{2.3153} & 1.6891  & \textbf{1.5918 } \\
          & MAE   & 1.0788  & \textbf{1.0773}  & 1.0591  & \textbf{1.0161}  & \textbf{0.8431}  & \textbf{0.8649 } \\
    \midrule
    \multirow{2}[2]{*}{Weather} & MSE   & 0.2835  & \textbf{0.2834 } & 0.3277  & \textbf{0.3112 } & \textbf{0.2792 } & 0.2817  \\
          & MAE   & 0.3324  & \textbf{0.3323 } & 0.3651  & \textbf{0.3589 } & 0.2983  & \textbf{0.2981 } \\
    \bottomrule
    \end{tabular}}
    \label{main table3}
\end{table}

\begin{table}[htbp!]
\centering
  \caption{Multivariate time series forecasting results on deep learning methods with/without hybrid loss framework (prediction
length is 720). }
  
\resizebox{0.92\textwidth}{!}{\begin{tabular}{cccccccc}
    \toprule
    \multirow{2}[4]{*}{Datasets} & Models & \multicolumn{2}{c}{Dlinear} & \multicolumn{2}{c}{FEDformer} & \multicolumn{2}{c}{Patchtst} \\
\cmidrule{2-8}          & Loss  & Original & Hybrid Loss & Original & Hybrid Loss & Original & Hybrid Loss \\
    \midrule
    \multirow{2}[2]{*}{ETTh1\newline{}} & MSE   & \textbf{0.5284}  & 0.5296 & \textbf{0.5022}  & 0.5053 & 0.4798 & \textbf{0.4778} \\
          & MAE   & \textbf{0.5185}  & 0.5193 & \textbf{0.5032}  & 0.5050 & 0.4707  & \textbf{0.4697} \\
    \midrule
    \multirow{2}[2]{*}{ETTh2\newline{}} & MSE   & 0.7736  & \textbf{0.7719} & 0.4804 & \textbf{0.4798} & 0.4118  & \textbf{0.4102 } \\
          & MAE   & 0.6313  & \textbf{0.6306} & 0.4873 & \textbf{0.4870} & 0.4334  & \textbf{0.4325 } \\
    \midrule
    \multirow{2}[2]{*}{ETTm1} & MSE   & 0.4794  & \textbf{0.4792 } & 0.5068  & \textbf{0.5065 } & 0.4556  & \textbf{0.4535} \\
          & MAE   & 0.4567  & \textbf{0.4566 } & \textbf{0.4887}  & \textbf{0.4887 } & \textbf{0.4359} & 0.4401  \\
    \midrule
    \multirow{2}[2]{*}{ETTm2} & MSE   & 0.4385 & \textbf{0.4383} & 0.4263  & \textbf{0.4252 } & 0.3986 & \textbf{0.3968 } \\
          & MAE   & 0.4439  & \textbf{0.4437 } & 0.4229  & \textbf{0.4221 } &0.3969  & \textbf{0.3954 } \\
    \midrule
    \multirow{2}[2]{*}{Electricity} & MSE   & \textbf{0.2425}  & \textbf{0.2425 } & 0.2639  & \textbf{0.2634}  & \textbf{0.2349 } & 0.2357  \\
          & MAE   & 0.3283  & \textbf{0.3283 } & \textbf{0.3669 } & 0.3664  & \textbf{0.3146 } & 0.3274  \\
    \midrule
    \multirow{2}[2]{*}{Exchange} & MSE   & 0.7816  & \textbf{0.7815 } & \textbf{1.1535 } & 1.1708  & \textbf{0.8363 } & 0.8411  \\
          & MAE   & 0.6692  & \textbf{0.6690 } & \textbf{0.8227 } & 0.8304  & \textbf{0.6879 } & 0.6898  \\
    \midrule
    \multirow{2}[2]{*}{illness} & MSE   & \textbf{2.5735}  & 2.5765  & 2.7804  & \textbf{2.5085 } & 1.6775  & \textbf{1.5200 } \\
          & MAE   & \textbf{1.1578 } & 1.1591   & 1.1323  & \textbf{1.1163} & 0.8754  & \textbf{0.8052 } \\
    \midrule
    \multirow{2}[2]{*}{Weather} & MSE   & 0.3484  & \textbf{0.3483 } & \textbf{0.3721}  & 0.3848  & \textbf{0.3514 } & 0.3557  \\
          & MAE   & 0.3849  & \textbf{0.3848 } & 0.4187  & \textbf{0.3589 } & \textbf{0.3461}  & 0.3473  \\
    \bottomrule
    \end{tabular}}
    \label{main table4}
\end{table}






\section{More Results of Ablation Study }
\label{More Results of Ablation Study}

Due to the high computational cost of the Fourier transform in FEDformer, the second ablation study as described in the main text is conducted only on the ETTh1 and ETTh2 datasets for FEDformer. Results are shown in Table \ref{abla3}. The results further corroborate the conclusions presented in the main paper, which confirm that \textbf{uniform initial weights (set to 0.5) is an effective initialization strategy, allowing the model to subsequently and efficiently adjust the individual loss weights}.

\begin{table}[htbp!]
  \centering
  \caption{The ablation study of multivariate time series forecasting results on our hybrid loss framework with different initial weights. As $w_1 + w_2 = 1$ and $\alpha + \beta = 1$, we only specify the initial values of $w_1$ and $\alpha$ in the table.}
    \resizebox{0.90\textwidth}{!}{\begin{tabular}{ccccccccccc}
    \toprule
    \multicolumn{1}{c}{} &       &       & \multicolumn{2}{c}{ETTh1\newline{}} & \multicolumn{2}{c}{ETTh2\newline{}} & \multicolumn{2}{c}{ETTm1} & \multicolumn{2}{c}{ETTm2} \\
    \midrule
    Models & $w_1$  & $\alpha$ & MSE   & MAE   & MSE   & MAE   & MSE   & MAE   & MSE   & MAE \\
    \midrule
    \multirow{5}[6]{*}{Dlinear} & \multirow{2}[2]{*}{0.1} & 0.1   & 0.4596 & 0.4524 & 0.4939 & 0.4779 & 0.4091 & 0.4142 & 0.3064 & 0.3638 \\
          &       & 0.9   & 0.4610 & 0.4541 & 0.4978 & 0.4791 & 0.4047 & 0.4097 & 0.3091 & 0.3658 \\
\cmidrule{2-11}          & 0.5   & 0.5   & 0.4579 & 0.4511 & 0.4974 & 0.4785 & 0.4060 & 0.4102 & 0.3100 & 0.3667 \\
\cmidrule{2-11}          & \multirow{2}[2]{*}{0.9} & 0.1   & 0.4589 & 0.4519 & 0.4982 & 0.4793 & 0.4050 & 0.4103 & 0.3102 & 0.3670 \\
          &       & 0.9   & 0.4589 & 0.4519 & 0.4983 & 0.4793 & 0.4050 & 0.4103 & 0.3102 & 0.3670 \\
    \midrule
    \multirow{5}[6]{*}{Patchtst} & \multirow{2}[2]{*}{0.1} & 0.1   & 0.4554 & 0.4451 & 0.3651 & 0.3940 & 0.3836 & 0.3975 & 0.2802 & 0.3200 \\
          &       & 0.9   & 0.4518 & 0.4412 & 0.3650 & 0.3937 & 0.3828 & 0.3973 & 0.2800 & 0.3254 \\
\cmidrule{2-11}          & 0.5   & 0.5   & 0.4502 & 0.4402 & 0.3639 & 0.3929 & 0.3813 & 0.3943 & 0.2790 & 0.3247 \\
\cmidrule{2-11}          & \multirow{2}[2]{*}{0.9} & 0.1   & 0.4493 & 0.4393 & 0.3642 & 0.3935 & 0.3821 & 0.3960 & 0.2792 & 0.3247 \\
          &       & 0.9   & 0.4492 & 0.4391 & 0.3642 & 0.3934 & 0.3821 & 0.3935 & 0.2792 & 0.3248 \\
    \midrule 
    \multirow{5}[6]{*}{FEDformer} & \multirow{2}[2]{*}{0.1} & 0.1   & 0.4428 & 0.4601 & 0.4432 & 0.4549 & --- & --- & --- & --- \\
          &       & 0.9   & 0.4421 & 0.4596 & 0.4441 & 0.4556 & --- & --- & --- & --- \\
\cmidrule{2-11}          & 0.5   & 0.5   & 0.4380 & 0.4573 & 0.4417 & 0.4539 & --- & ---  & --- & --- \\
\cmidrule{2-11}          & \multirow{2}[2]{*}{0.9} & 0.1   & 0.4394 & 0.4581 & 0.4430 & 0.4548 & --- & ---& --- & --- \\
          &       & 0.9   & 0.4394 & 0.4581 & 0.4430& 	0.4549 & --- & --- & --- & ---\\
    
    \bottomrule
    \end{tabular}}
  \label{abla3}%
\end{table}%

\section{More Showcases}
\label{More Showcases}
\begin{figure*}[htb!]
    \centering
    \subfloat[FEDformer on ETTh1.]{\includegraphics[scale=0.30]{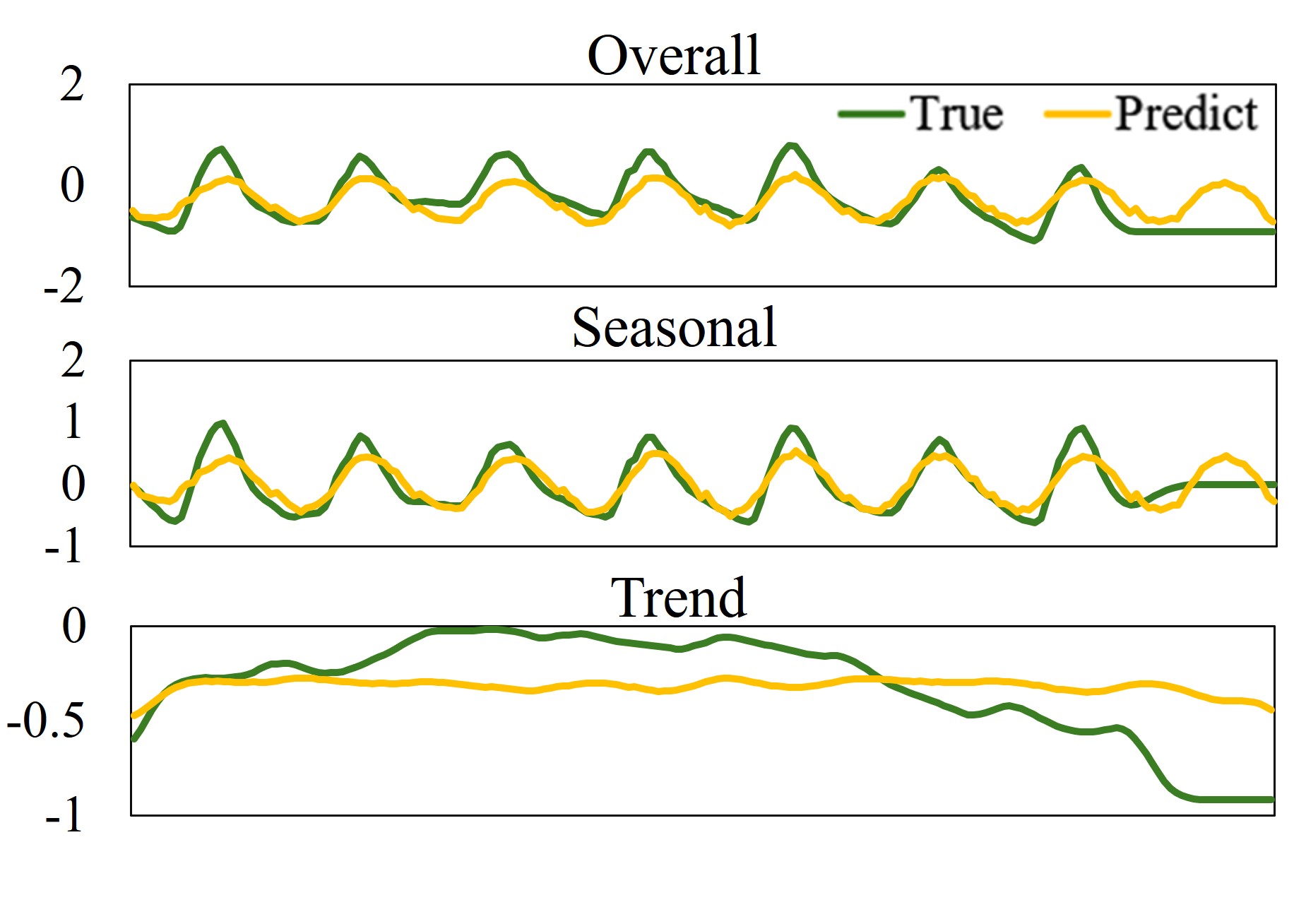}}
    \quad
    \subfloat[Patchtst on ETTh2.]{\includegraphics[scale=0.30]{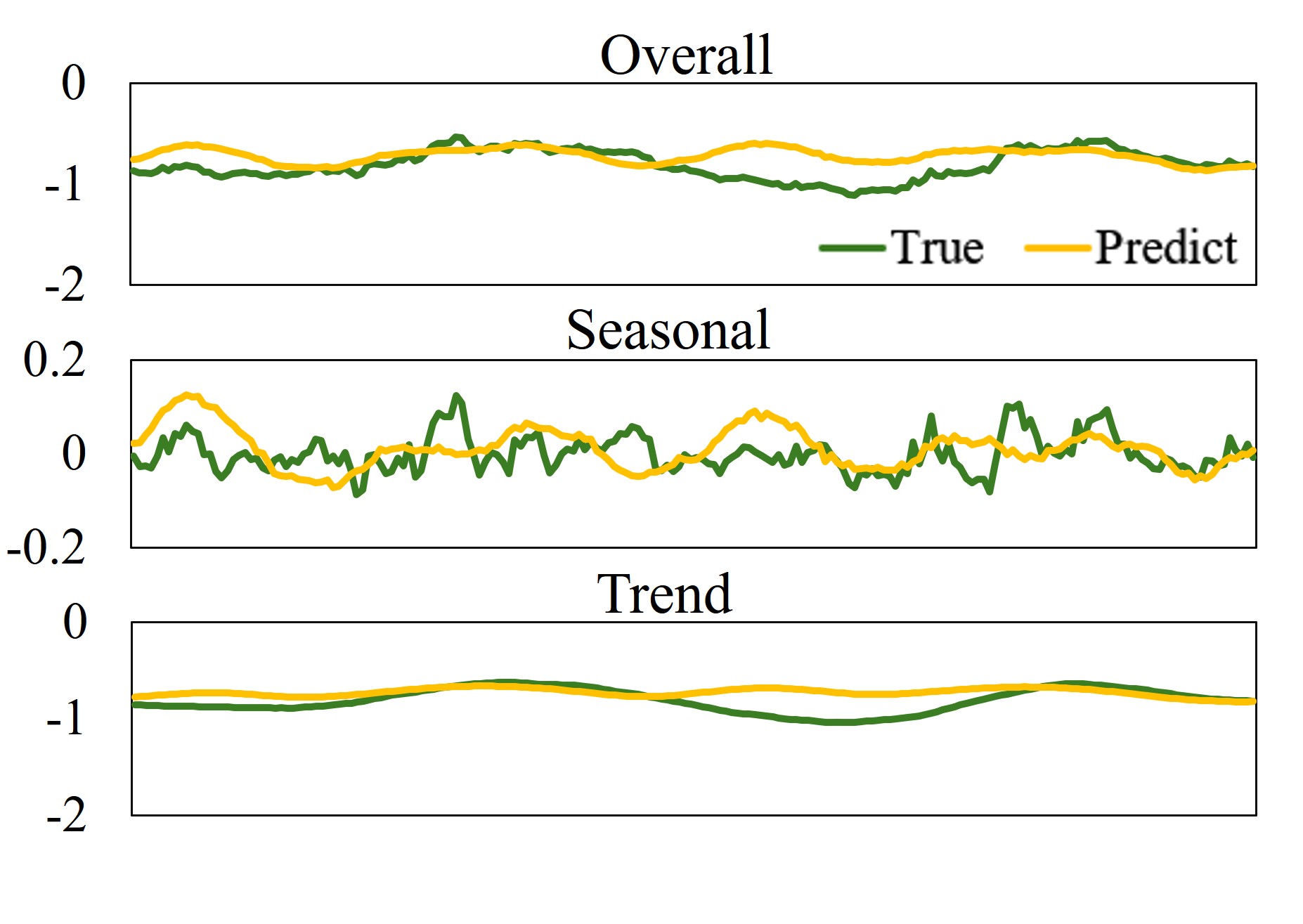}}
    \quad
    \subfloat[Dlinear on ETTm1.]{\includegraphics[scale=0.30]{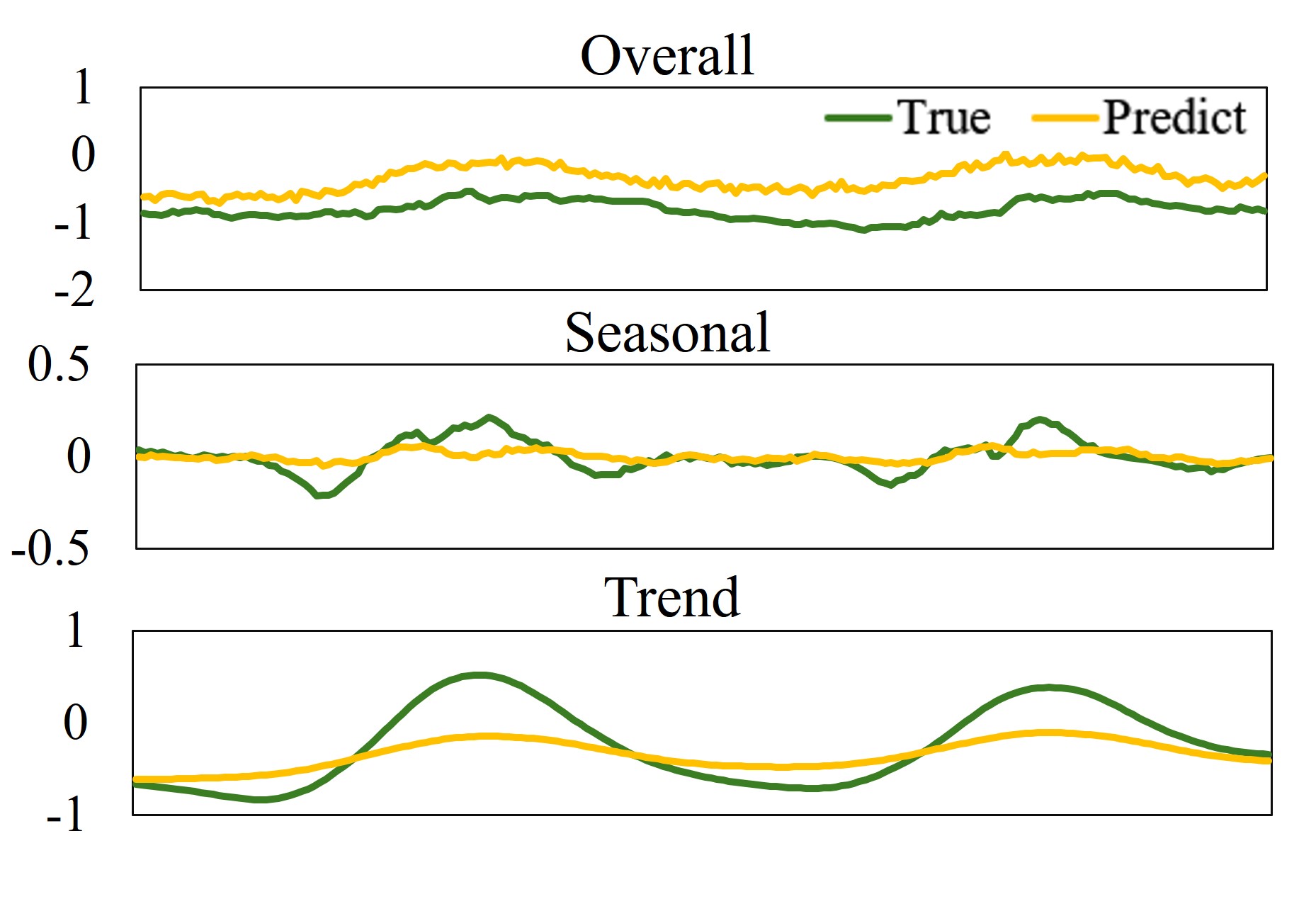}}
    \newline
    \subfloat[FEDformer with our hybrid loss framework on ETTh1.]{\includegraphics[scale=0.30]{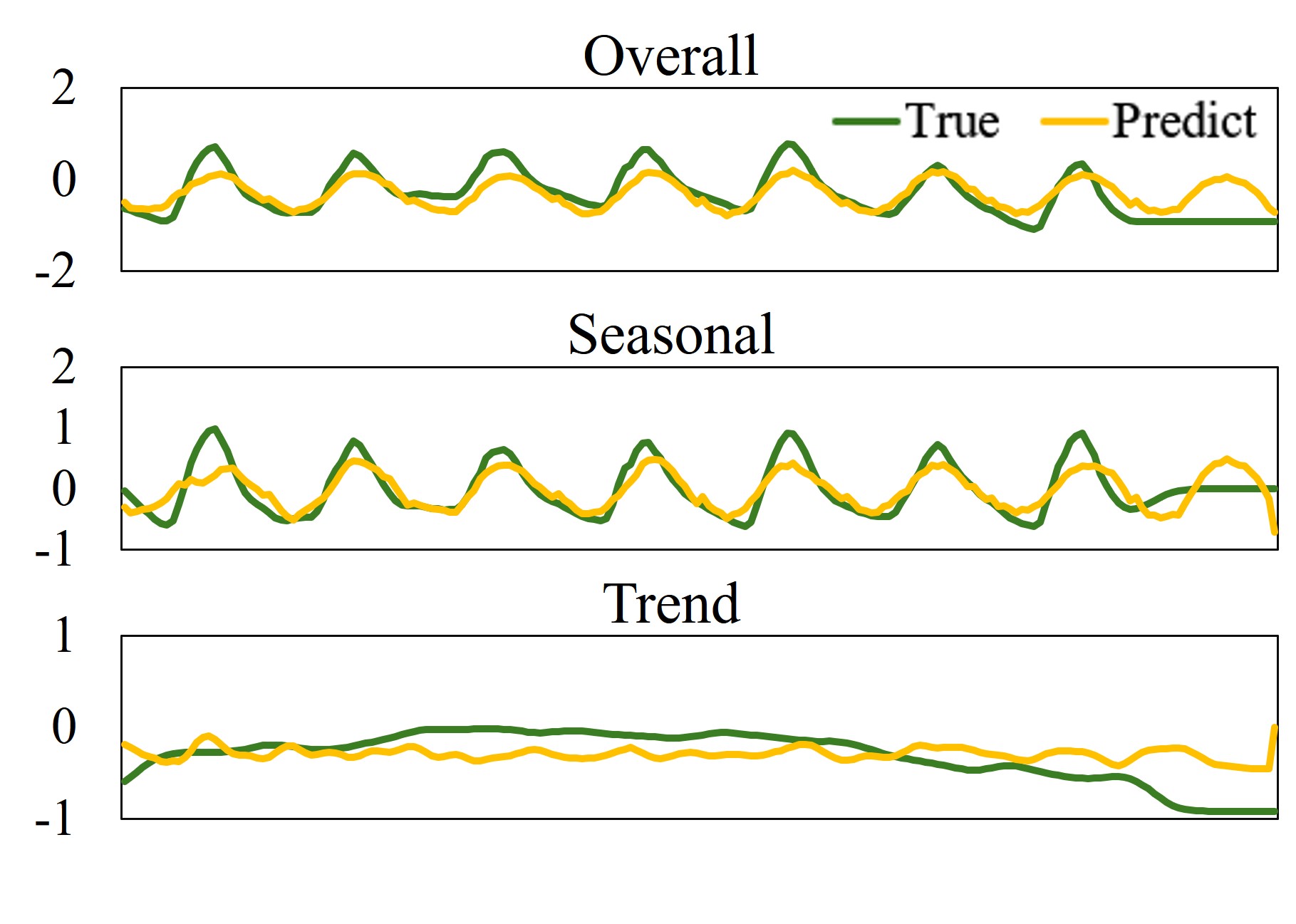}}
    \quad
    \subfloat[Patchtst with our hybrid loss framework on ETTh2.]{\includegraphics[scale=0.30]{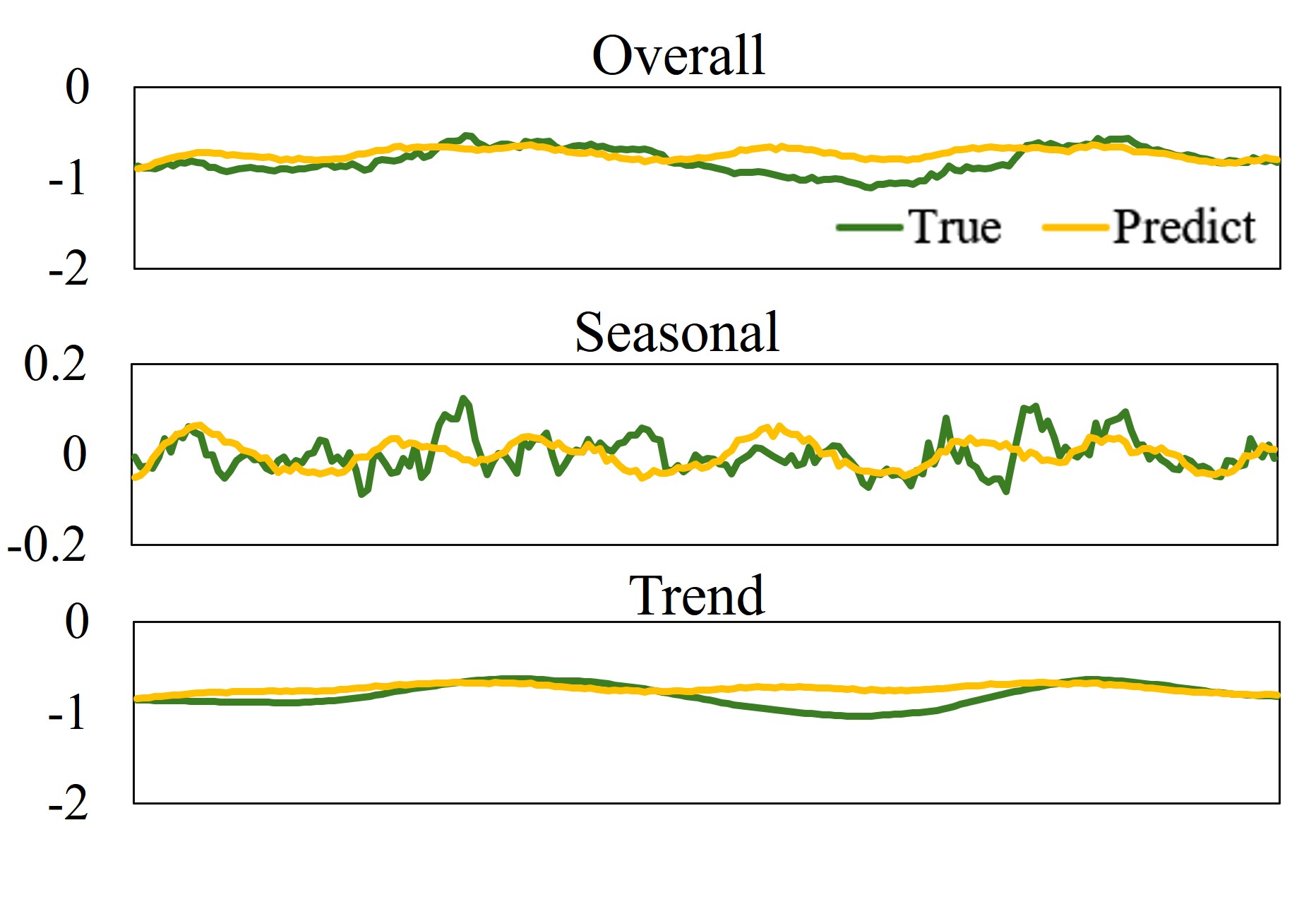}}
    \quad
    \subfloat[Dlinear  with our hybrid loss framework on ETTm1.]{\includegraphics[scale=0.30]{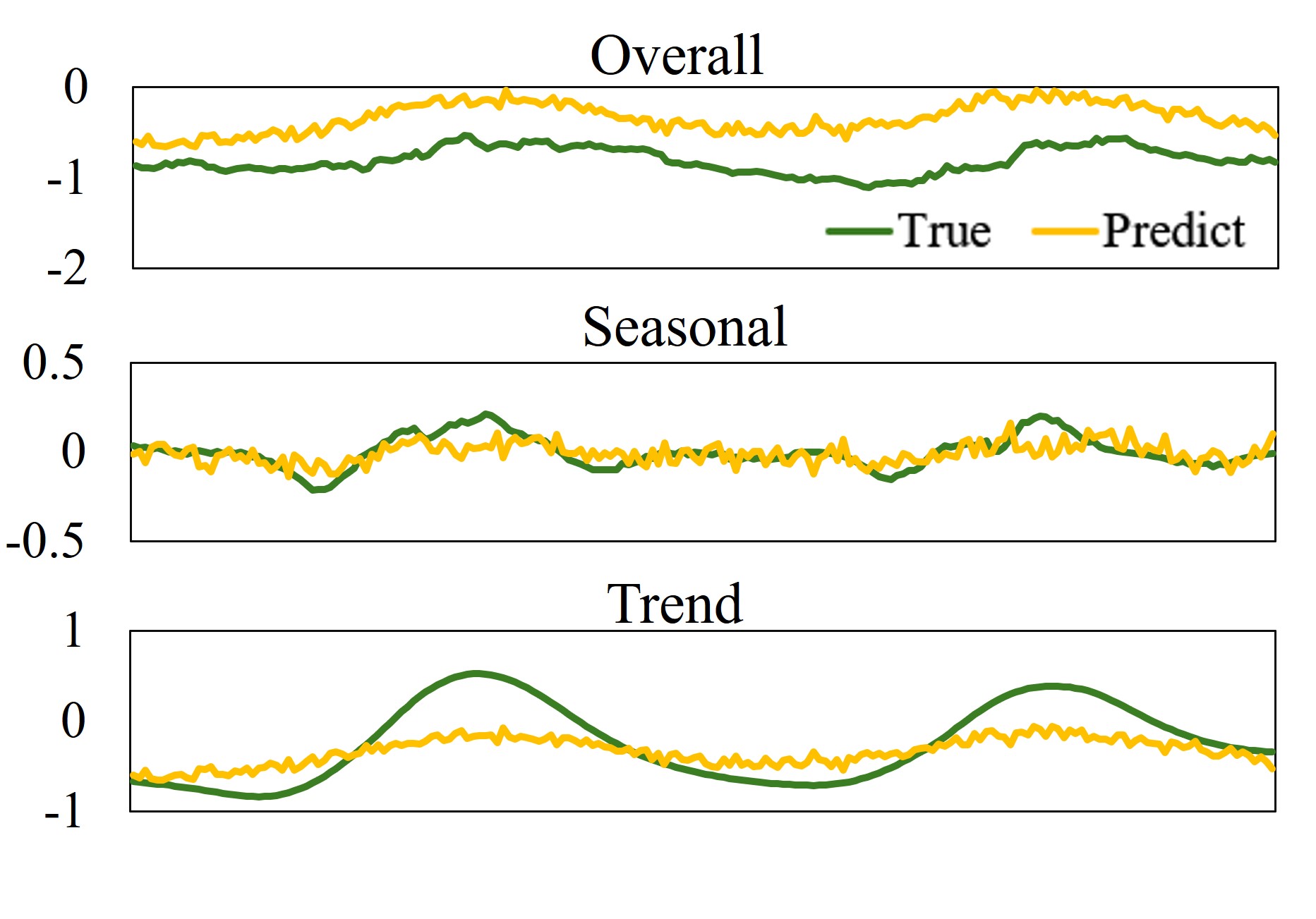}}
    \caption{The case study of time series forecasting. The results show the prediction-length-192 part (input length is 96) for different methods on different datasets. Each sub figure presents the single-variate (last variate) overall forecasting part and the forecasting part of the individual sub-series.}
    \label{fig:case-192}
\end{figure*}

\begin{figure*}[htb!]
    \centering
    \subfloat[FEDformer on ETTh1.]{\includegraphics[scale=0.30]{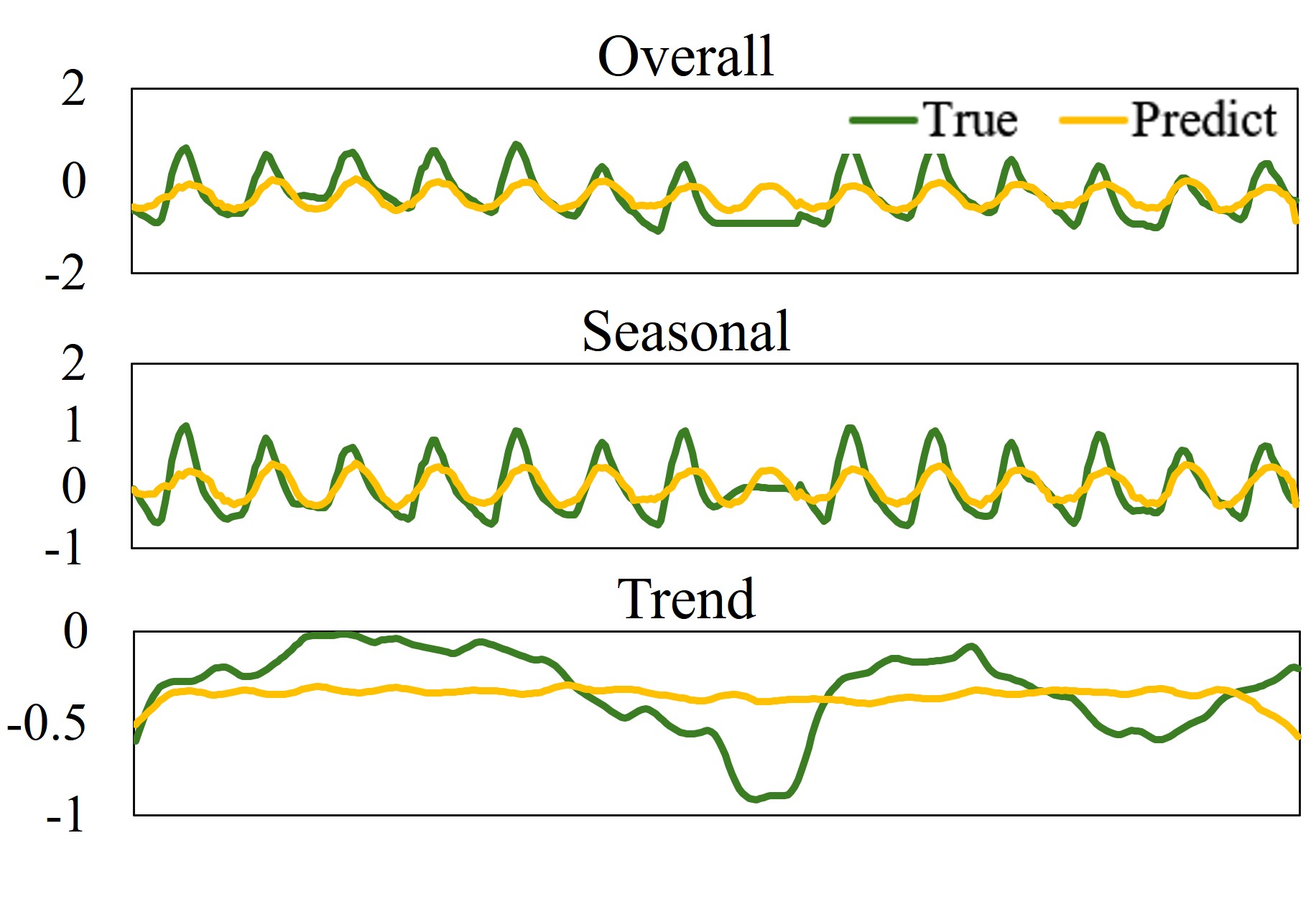}}
    \quad
    \subfloat[Patchtst on ETTh2.]{\includegraphics[scale=0.30]{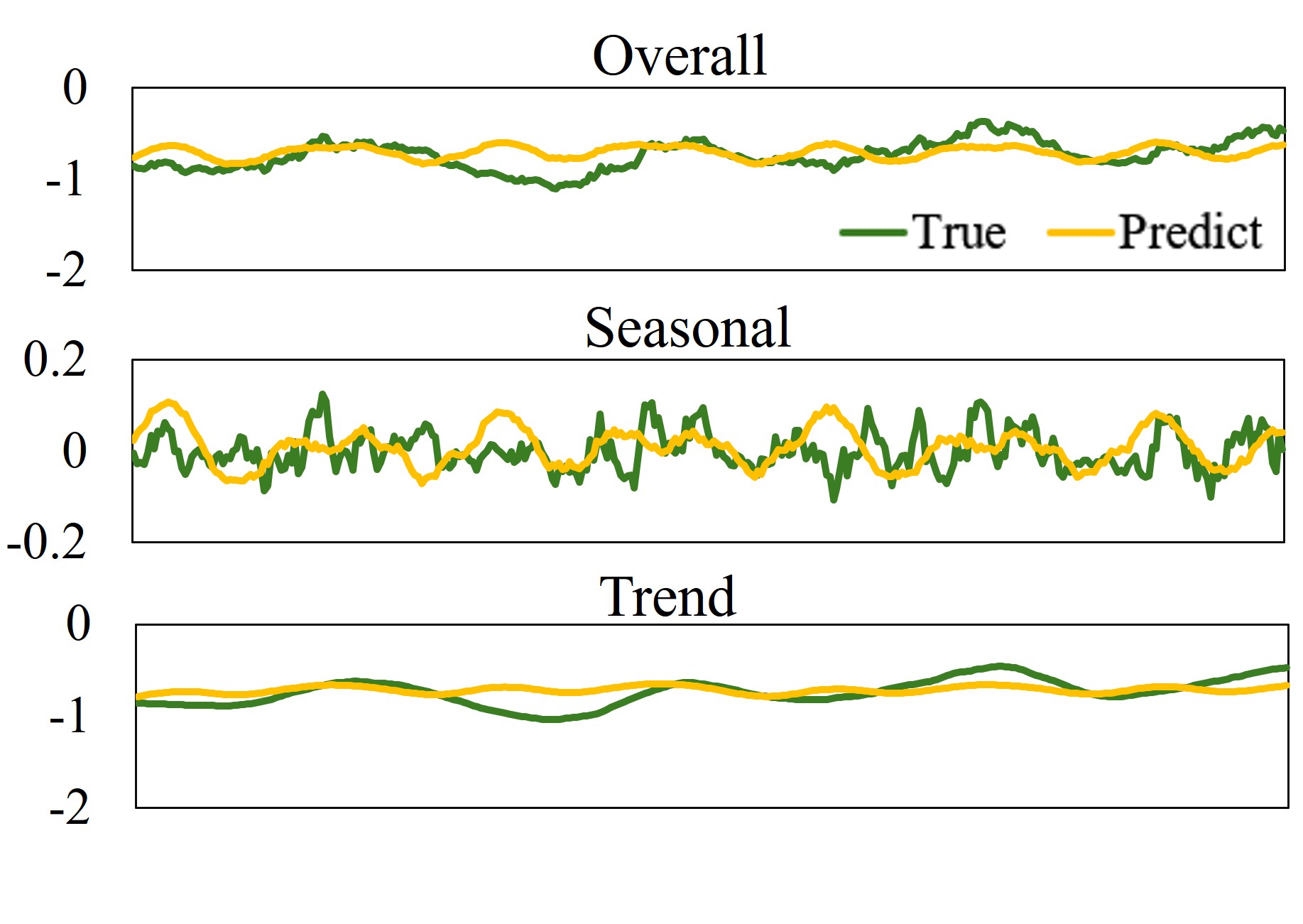}}
    \quad
    \subfloat[Dlinear on ETTm1.]{\includegraphics[scale=0.30]{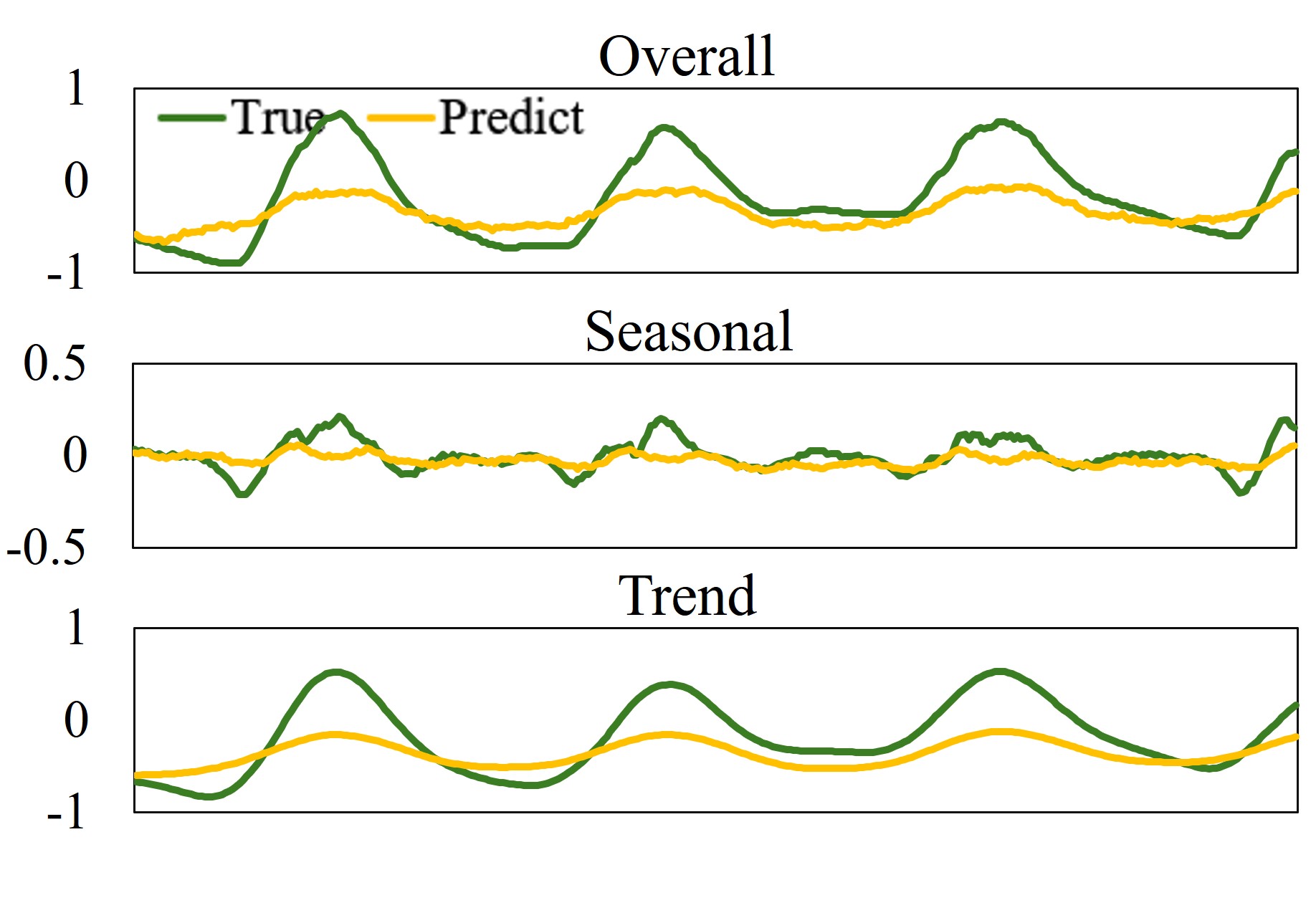}}
    \newline
    \subfloat[FEDformer with our hybrid loss framework on ETTh1.]{\includegraphics[scale=0.30]{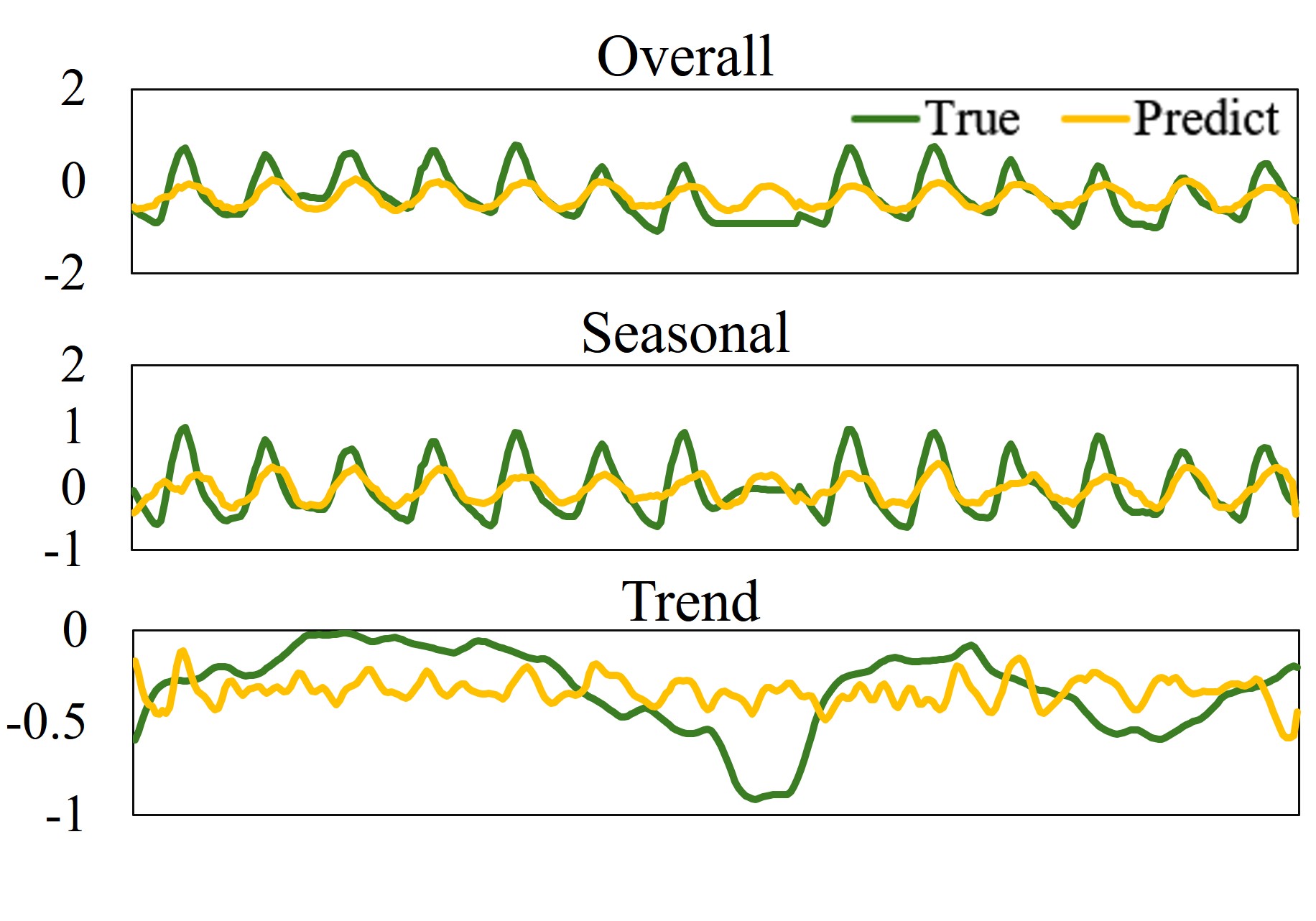}}
    \quad
    \subfloat[Patchtst with our hybrid loss framework on ETTh2.]{\includegraphics[scale=0.30]{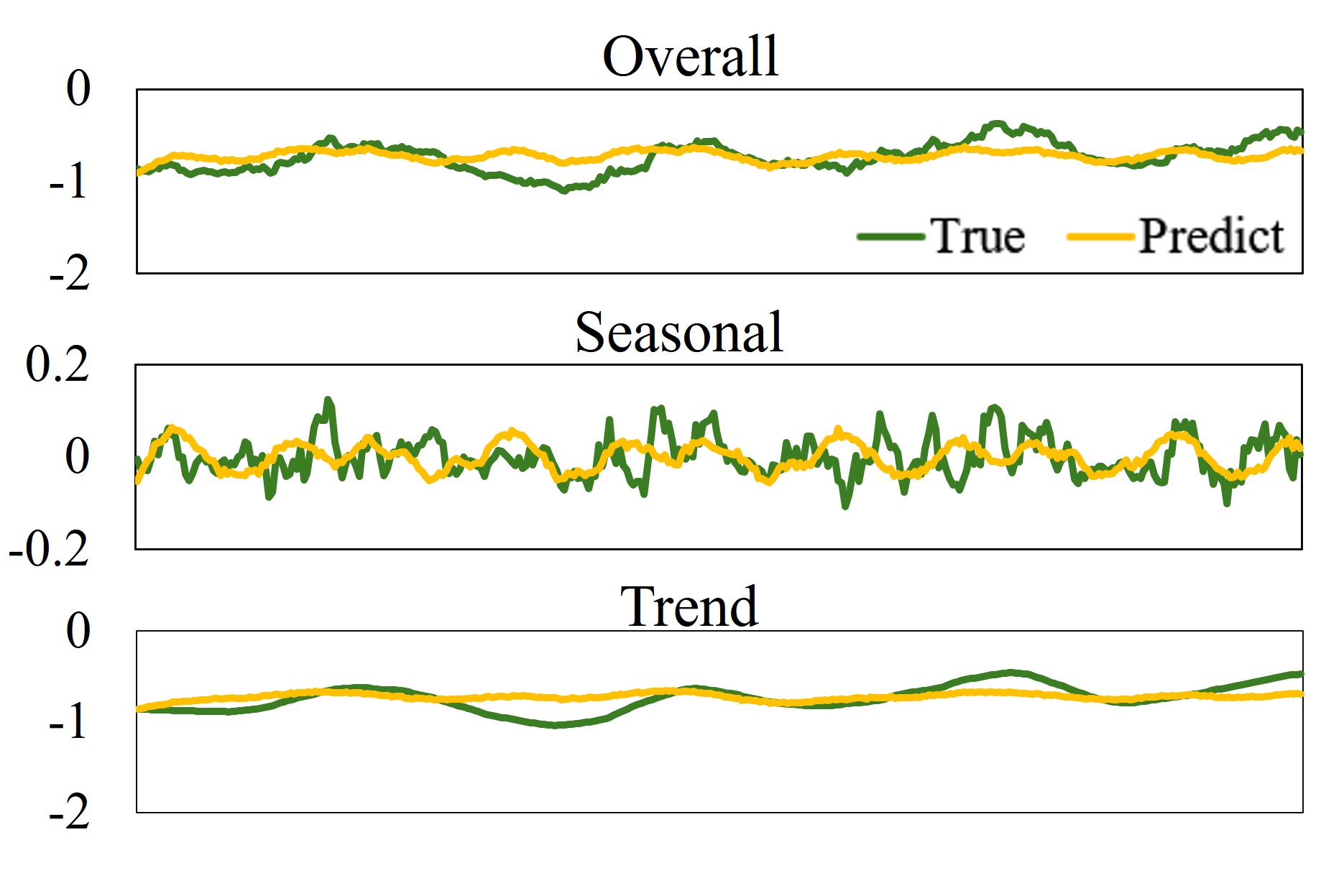}}
    \quad
    \subfloat[Dlinear  with our hybrid loss framework on ETTm1.]{\includegraphics[scale=0.30]{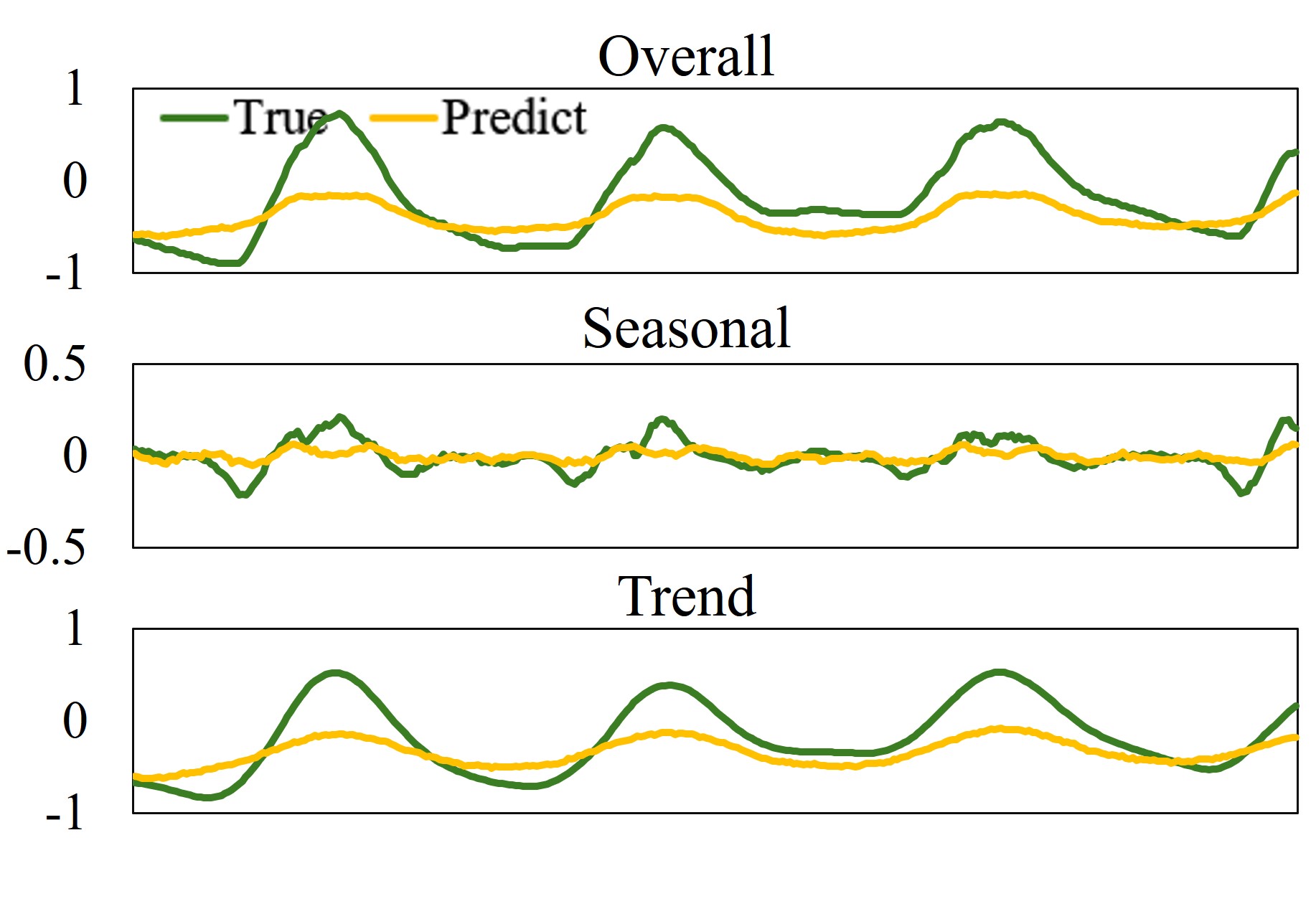}}
    \caption{The case study of time series forecasting. The results show the prediction-length-336 part (input length is 96) for different methods on different datasets. Each sub figure presents the single-variate (last variate) overall forecasting part and the forecasting part of the individual sub-series.}
    \label{fig:case-336}
\end{figure*}

\begin{figure*}[htb!]
    \centering
    \subfloat[FEDformer on ETTh1.]{\includegraphics[scale=0.30]{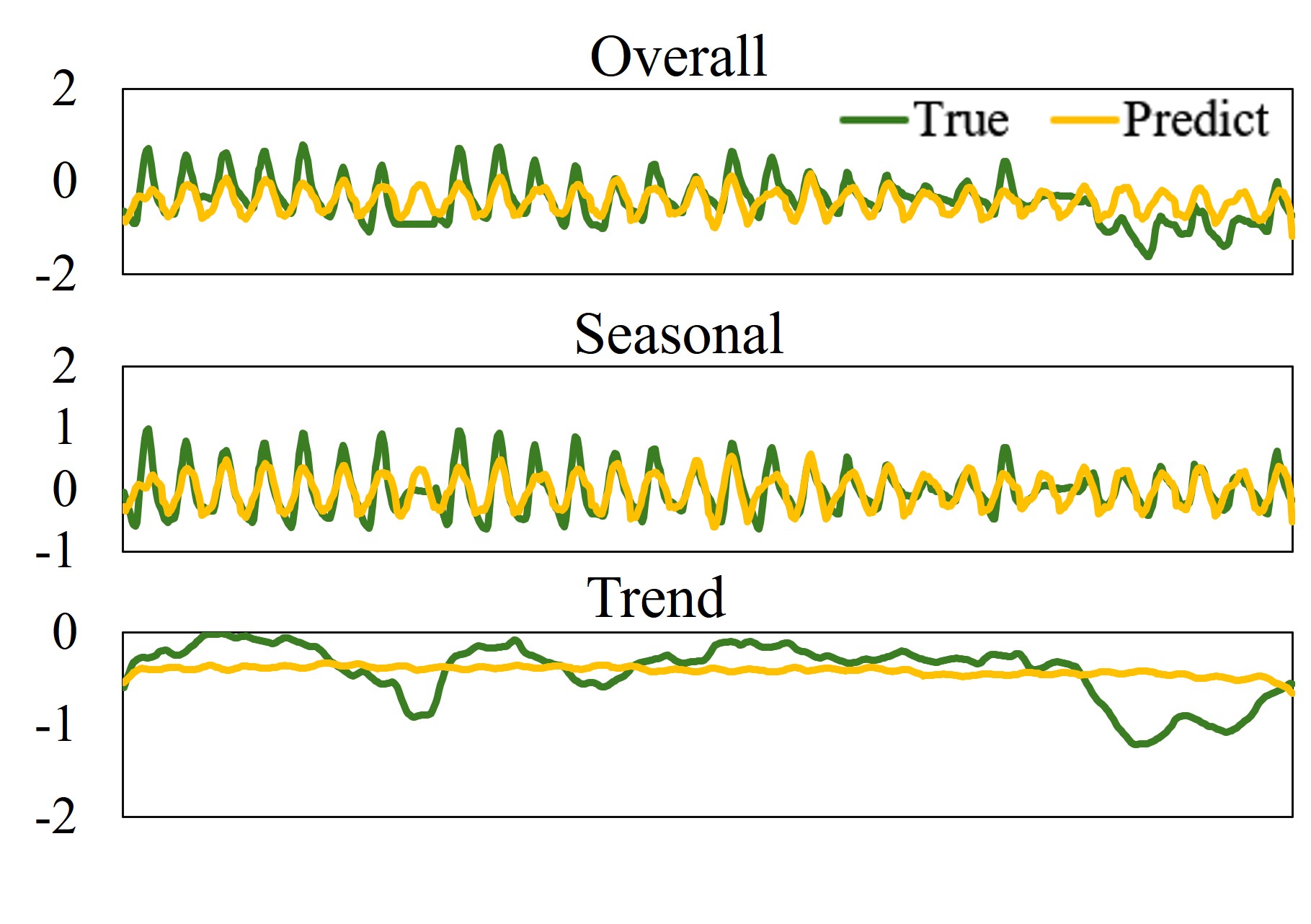}}
    \quad
    \subfloat[Patchtst on ETTh2.]{\includegraphics[scale=0.30]{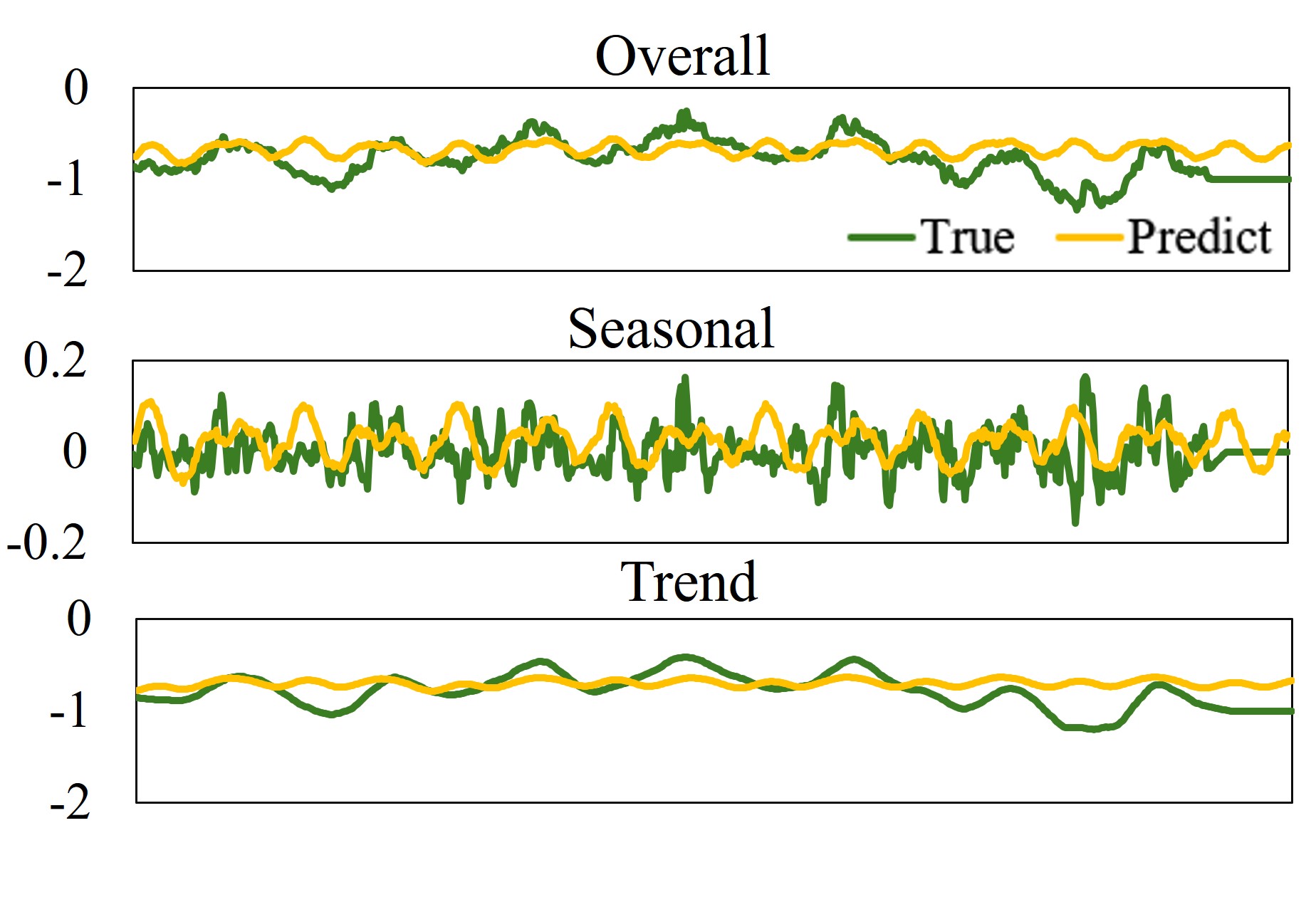}}
    \quad
    \subfloat[Dlinear on ETTm1.]{\includegraphics[scale=0.30]{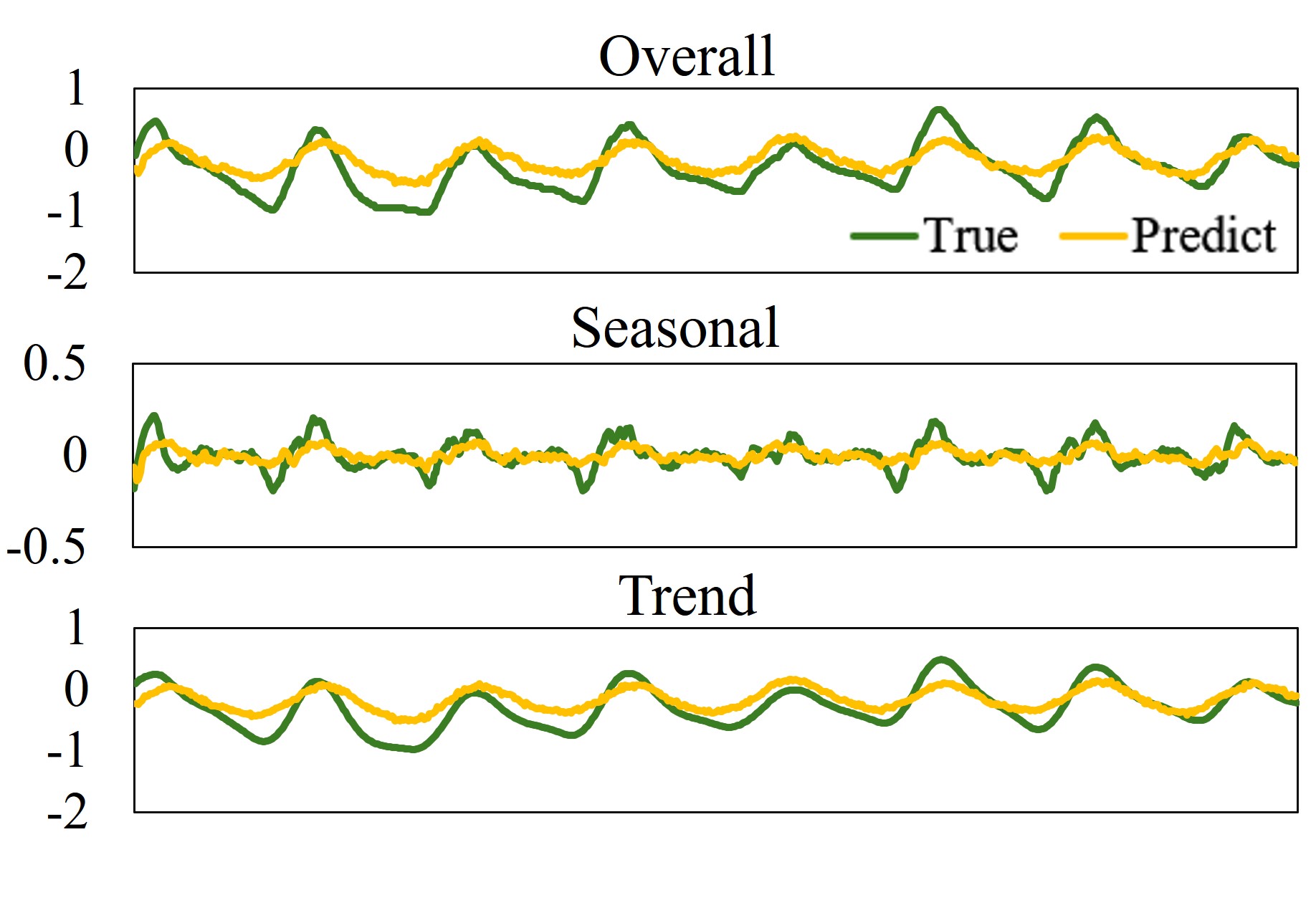}}
    \newline
    \subfloat[FEDformer with our hybrid loss framework on ETTh1.]{\includegraphics[scale=0.30]{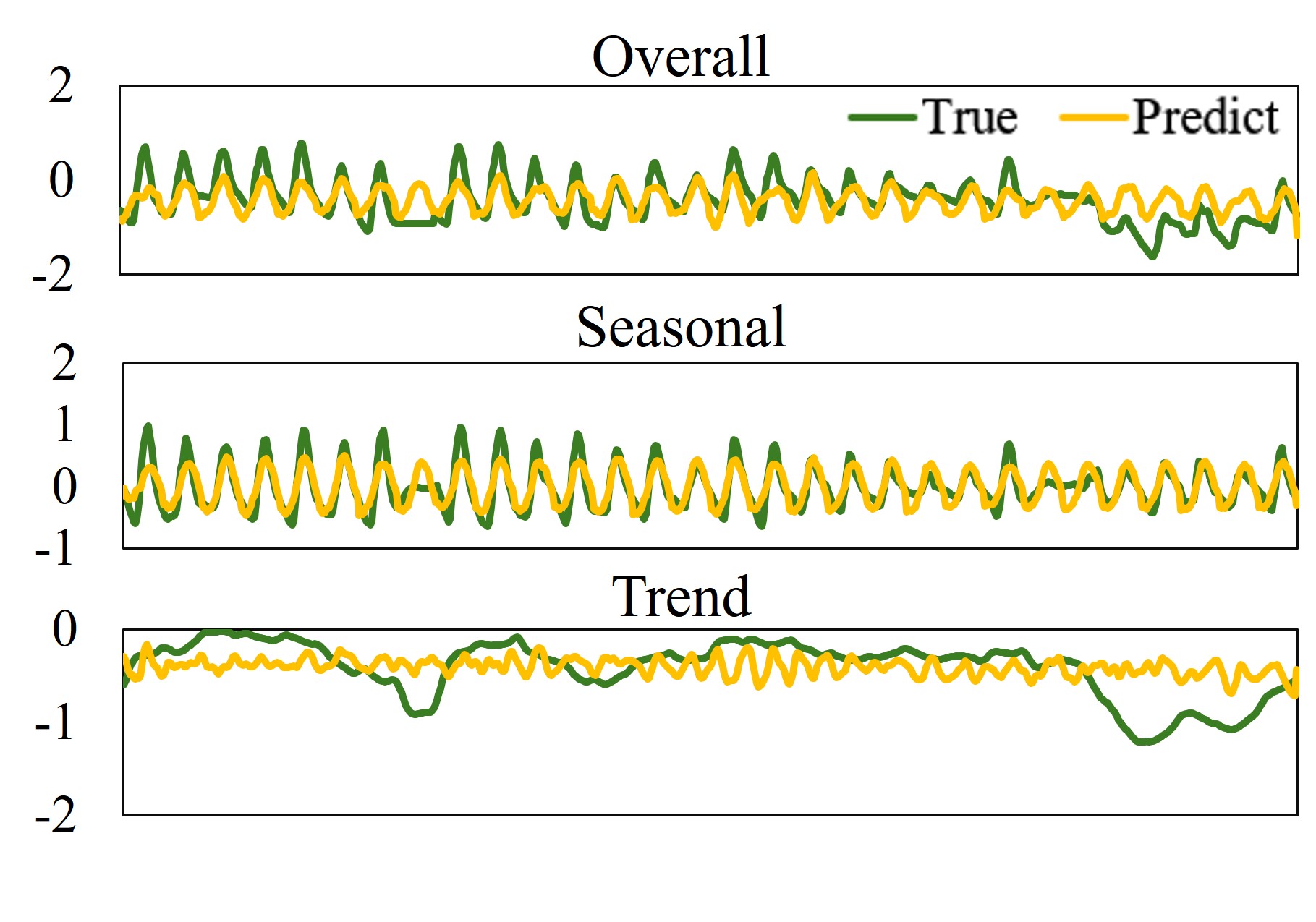}}
    \quad
    \subfloat[Patchtst with our hybrid loss framework on ETTh2.]{\includegraphics[scale=0.30]{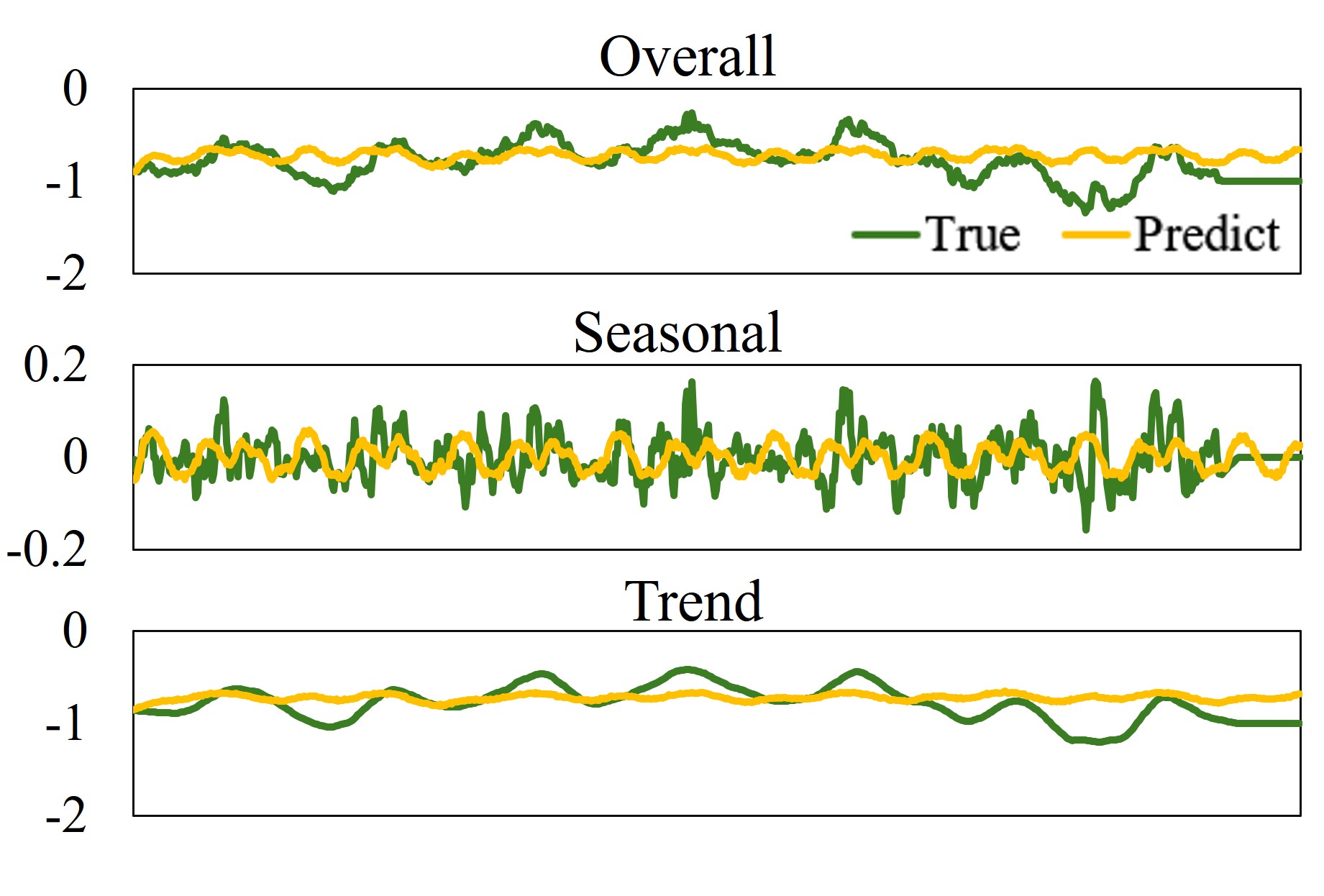}}
    \quad
    \subfloat[Dlinear  with our hybrid loss framework on ETTm1.]{\includegraphics[scale=0.30]{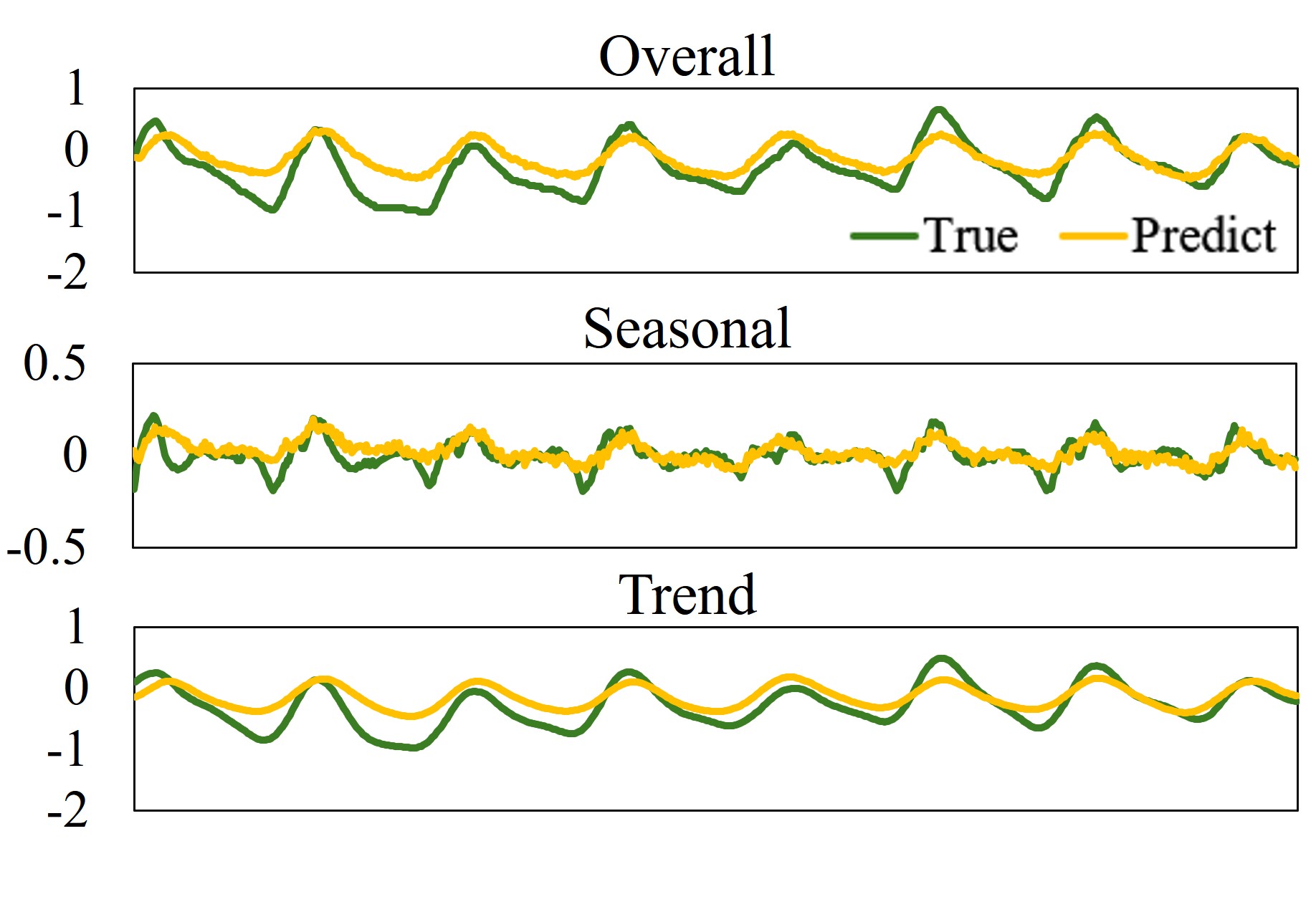}}
    \caption{The case study of time series forecasting. The results show the prediction-length-720 part (input length is 96) for different methods on different datasets. Each sub figure presents the single-variate (last variate) overall forecasting part and the forecasting part of the individual sub-series.}
    \label{fig:case-720}
\end{figure*}

This section presents additional cases using the original overall loss and our hybrid loss framework, as illustrated in Figure \ref{fig:case-192}, Figure \ref{fig:case-336}, and Figure \ref{fig:case-720}. Notably, we show the results of the forecasting part with the settings of the input length 96 and prediction length 96 in the main text. Therefore, we show the results of the forecasting part with the settings of the input length 96 and prediction length \{192, 336, 720\} here, respectively. These results further support our conclusions from the main text: \textbf{the original overall loss may lead to large errors in individual sub-series, further hindering overall forecasting performance, and our hybrid loss framework effectively mitigates this issue.}

\end{document}